\documentclass{article} %
\PassOptionsToPackage{table}{xcolor}
\usepackage{iclr2027_conference,times}

\usepackage{amsmath,amsfonts,bm}

\def\eqref#1{equation~\ref{#1}}

\def\1{\bm{1}}

\DeclareMathAlphabet{\mathsfit}{\encodingdefault}{\sfdefault}{m}{sl}
\SetMathAlphabet{\mathsfit}{bold}{\encodingdefault}{\sfdefault}{bx}{n}

\usepackage{hyperref}
\usepackage{url}

\usepackage{graphicx}
\usepackage{booktabs} 

\usepackage[utf8]{inputenc} %
\usepackage[T1]{fontenc}    %
\usepackage{hyperref}       %
\usepackage{url}            %
\usepackage{amsfonts}       %
\usepackage{nicefrac}       %
\usepackage{microtype}      %
\usepackage{tabularx}
\usepackage{rotating}
\usepackage{makecell}
\usepackage{pifont}
\usepackage[table]{xcolor}
\usepackage{array}
\usepackage{amsmath}

\newcommand{\na}{---}
\newcommand{\good}[1]{\cellcolor{cellgood}#1}
\newcommand{\midc}[1]{\cellcolor{cellmid}#1}
\newcommand{\bad}[1]{\cellcolor{cellbad}#1}
\newcommand{\ourscell}[1]{\cellcolor{cellours}\textbf{#1}}

\definecolor{cellgood}{RGB}{220,245,225}
\definecolor{cellmid}{RGB}{255,242,204}
\definecolor{cellbad}{RGB}{248,218,218}
\definecolor{cellours}{RGB}{218,232,252}

\title{REALIS: A Curated Dataset for Studying the Challenges of AI Image Detection}

\author{%
Aleksandr Gushchin$^{1,2}$, Khaled Abud$^{1,2}$, Georgii Bychkov$^{2,1}$, Ekaterina Shumitskaya$^{2,1}$, \\
\bf Artem Filippov$^{3}$, Sergey Lavrushkin$^{1,2}$, Dmitriy S.~Vatolin$^{1}$, Anastasia Antsiferova$^{1,2}$ \\[1ex]
\normalfont $^{1}$MSU Institute for Artificial Intelligence \quad
$^{2}$Trusted AI Research Center RAS \\
\normalfont $^{3}$Lomonosov Moscow State University}

\iclrfinalcopy %
\begin{document}

\maketitle
\lhead{Preprint}

\begin{abstract}
AI-generated image detectors are often evaluated on benchmarks where real and synthetic images differ in content, quality, or generation artifacts, allowing models to rely on dataset-specific cues and fail on unfamiliar generators or processed images. Existing datasets provide limited support for evaluating these challenges jointly across diverse visual content. We introduce REALIS, a dataset of 1.43 million real and synthetic images generated by 42 modern text-to-image models, including the latest proprietary systems such as Nano Banana 2. REALIS combines prompts derived from real images, quality filtering, and stratified sampling to reduce class-specific shortcuts while preserving content diversity. We further introduce REALIS-Expert, a stress-test subset for high-quality synthetic images, where real and generated samples are selected with closely matched semantic and visual characteristics. We also propose a robustness protocol covering 35 transformations at five severity levels to analyze detector behavior under image processing. Based on REALIS, our benchmark evaluates pretrained detectors, fine-tuned models, and zero-shot vision-language models under generator and post-processing shifts. On the hardest processed split, the best pretrained conventional detector achieves 0.550 ROC-AUC, compared with 0.752 for the best REALIS-trained detector. REALIS provides a unified framework for measuring and improving the reliability of AI-image detectors under conditions that better reflect real-world use.
\end{abstract}

\section{Introduction}

AI-generated images are becoming increasingly difficult to distinguish from photographs as text-to-image (T2I) systems improve. Reliably detecting AI-generated images remains crucial for critical applications like media forensics and content moderation. A robust detector must remain reliable when the generator is unfamiliar, the image content differs from its training distribution, and the image has undergone the processing common to online media.

Current benchmarks often make this reliability hard to measure. Their structure may be unsuitable for evaluating detector performance because of inherited biases and limited distortion methods. Those experiments may reflect recognition of dataset-specific cues rather than generation traces~(\cite{grommelt2024fakejpeg,rajan2024aligned}), and they may degrade performance on generators not encountered during training~(\cite{ojha2023towards,ren2026outofthebox}). Finally, these benchmarks test robustness only for simple processing techniques like resizing, cropping, and JPEG compression. More modern postprocessing applications, like neural image compression or watermarking, could suppress forensic traces or create new artifacts that trigger false positives~(\cite{wang2020cnn,corvi2023dm}). Strong performance on a clean, content-misaligned benchmark therefore need not
translate into reliable real-world behavior.

We believe that these problems should be solved jointly in one dataset. Recent datasets and benchmarks have expanded coverage along several of the discussed axes ~(\cite{zhu2023genimage,hong2025wildfake,pellegrini2025aigenbench,gushchin2026ntire}), but they lack support for content alignment and generator biases, and they use only simple post-processing strategies within a single controlled framework.
To address these issues, we introduce \textsc{Realis}, a dataset and benchmark designed around these three requirements. It contains 1.43 million real and fake images spanning 42 modern T2I generators. Each fake image is generated from a prompt derived from a real image, aligning data distributions of real and fake parts of the dataset. We combine strict real-image curation, symmetric filtering of both parts, and prompt-disjoint splits to reduce class-specific biases. We sample validation and test sets to ensure progressively increasing difficulty. We also introduce a post-processing protocol comprising 35 different transformations in eleven groups and five severity levels.

Overall, our contributions are:

\begin{itemize}
    \item We introduce \textsc{Realis}, a 1.43M-image dataset covering 42
    contemporary T2I generators. We use a specific filtering pipeline including  content-derived prompting, symmetric
    filtering, stratified sampling to reduce
    dataset-specific biases while preserving semantic diversity. We provide a test part of the dataset in the supplementary material  \url{https://anonymous-hf.com/a/89q3tdboxa2m/}.

    \item We present a protocol for testing the robustness of the evaluations to 35 modern transformations, including neural network-based ones, compound processing chains, and transformations
    held out from detector training.

    \item We benchmark 35 different detectors, including vision-language models. Controlled experiments show that
    even mild processing causes a sharp performance loss and that degradation
    is driven largely by increased false positives on real images.
\end{itemize}

\section{Related Work}

\paragraph{Datasets and Cross-Generator Generalization.}
ForenSynths~\cite{wang2020cnnspot} established an early cross-generator
evaluation protocol using GAN-based synthesis methods. GenImage
~\cite{zhu2023genimage} extended this setting to approximately 2.7M images
from eight generators and introduced cross-generator and degraded-image
evaluation. WildFake~\cite{hong2025wildfake} broadened the content and
generator distributions, while AI-GenBench
~\cite{pellegrini2025aigenbench} evaluates generalization to generators
released after those used for training. Community Forensics
~\cite{park2025community} instead maximizes generator diversity, collecting
2.7M images from 4,803 models and demonstrating that detector generalization
improves with the number and diversity of training generators. However, raw
generator count alone does not provide a controlled evaluation across modern
architectural families, generation-quality levels, and proprietary systems.
REALIS addresses this gap by curating 42 contemporary open and proprietary
generators and assigning them to generator-aware splits that measure
generalization to both seen and unseen generator families.

\paragraph{Robustness to Image Processing.}
Several benchmarks evaluate detectors beyond clean generator outputs.
GenImage studies basic blur, compression, and resolution changes, while
TrueFake~\cite{dellanna2025truefake} and RRDataset
~\cite{li2025rrbench} consider social-platform transmission, challenging
content, and re-digitization. The NTIRE 2026 challenge
~\cite{gushchin2026ntire} provides the closest setting to ours, combining 42
generators with 36 transformations and evaluating detectors on a mixture of
transformed and untransformed images. However, these protocols provide limited
support for isolating distortion severity and the cumulative effects of
multi-stage processing on content-aligned classes. REALIS addresses this
limitation using 35 transformations organized into eleven groups, stochastic
transformation chains at five severity levels, and paired evaluation of clean
and processed versions of the same source images.

Table~\ref{tab:dataset_comparison} compares our proposed dataset with the previous AI-generated image detection datasets and benchmarks. Our dataset combines general-content coverage, recent generators, easy access, and extensive robustness evaluation. Existing benchmarks emphasize particular aspects, such as generator breadth, or temporal generalization: our benchmark combines these dimensions together.

\section{Methodology}

\subsection{Dataset Construction}
\label{sec:dataset_construction}

\begin{figure*}[tbp]
    \centerline{\includegraphics[width=0.99\textwidth]{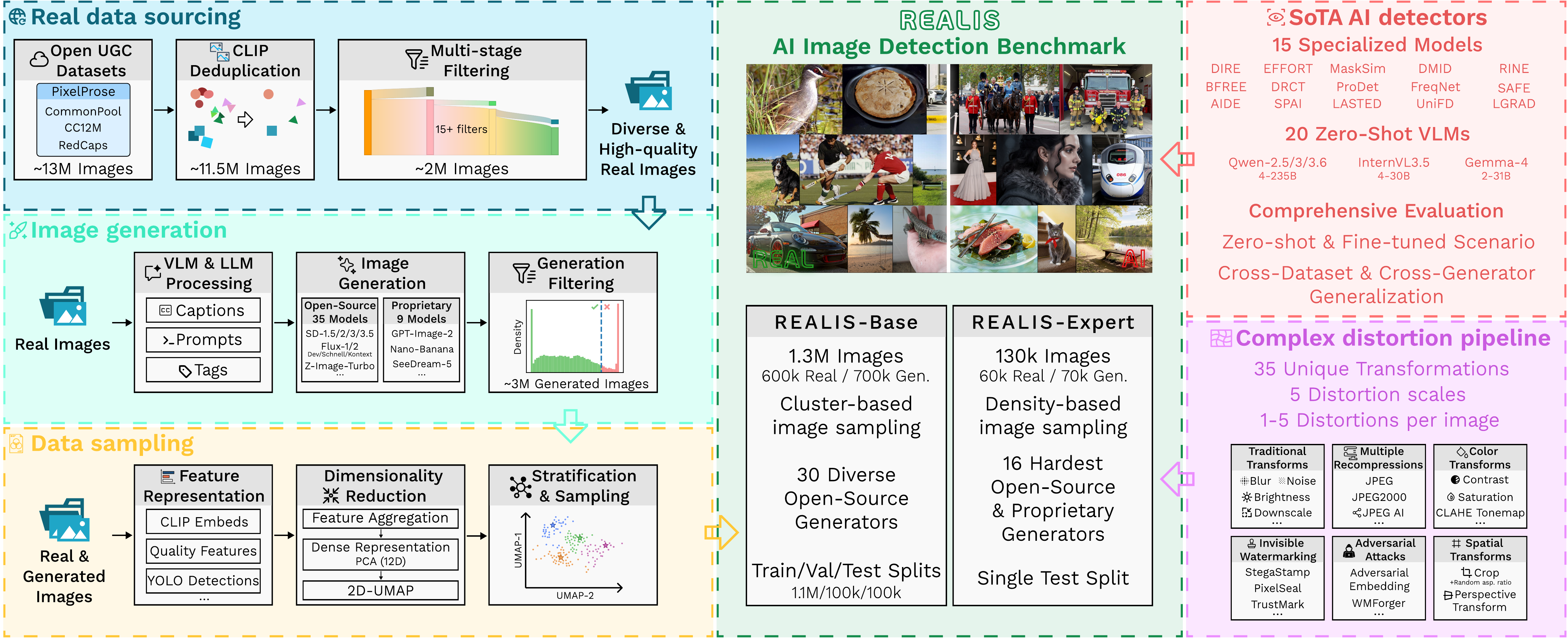}}
    \caption{Overview of REALIS construction and evaluation.}
\label{fig:overview}
\vspace{-10pt}
\end{figure*}

Figure~\ref{fig:overview} provides an overview of the construction pipeline.
It proceeds in four stages: sourcing and filtering a large pool of real
images; synthesizing a generation prompt for every retained real image and
rendering it with several T2I models; sampling the Train/Val/Test/Expert splits
from the resulting real--generated pool under two complementary criteria; and
finally passing the evaluation splits through a chained image degradation
pipeline. More details in App.\ref{app:dataset}

\subsubsection{Real Image Sourcing}

\paragraph{Source corpus.}
We source real images from PixelProse,~\cite{singla2024pixels} which aggregates CommonPool~\cite{gadre2023datacomp}, CC12M~\cite{changpinyo2021conceptual}, and RedCaps~\cite{desai2021redcaps} and provides dense captions and initial safety filtering. We apply additional filtering to remove low-resolution, watermarked, and duplicated images that could introduce class-specific shortcuts.

\begin{figure*}[tbp]
    \centerline{\includegraphics[width=0.99\textwidth]{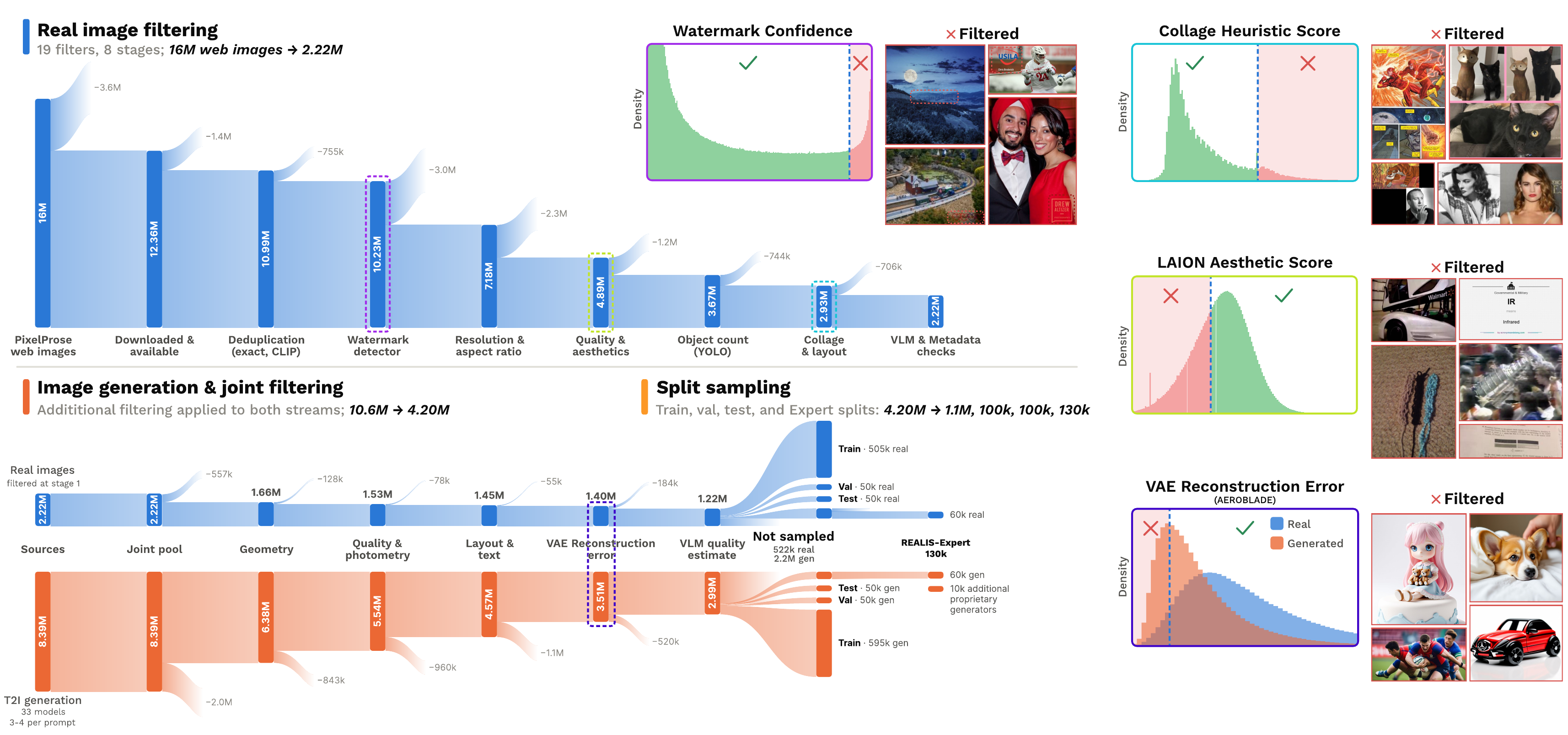}}
    \caption{\textbf{Left}: Filtering and sampling pipeline for REALIS. Flow widths represent image counts. \textbf{Right}: Highlighted distributions illustrate selected filtering criteria, with dashed lines marking thresholds and adjacent examples showing rejected images.}
\label{fig:filter_stats}
\vspace{-10pt}
\end{figure*}

\paragraph{Filtering cascade.}
Starting from $13$M candidate images, we apply a large collection of filters organized into five families. \emph{(i) Deduplication:} exact-duplicate removal on the joint feature signature, followed by near-duplicate removal in both the CLIP image and text embedding spaces. \emph{(ii) Low-level geometry and quality:} images must have reasonable resolution and aspect ratio (specifically, $\geq\!256$\,px at a short side and $\geq\!512$\,px at a long side, with an aspect ratio in $[0.33, 3]$), and must pass minimal image quality and aesthetic thresholds (as measured by TOPIQ~\cite{chen2024topiq} and LAION-Aesthetics models), set based on the distributions of the generated part of other tested AI-detection benchmarks, so that the real class is never systematically sharper or more attractive than the generated one --- a bias that would give detectors a shortcut unrelated to synthesis artifacts. \emph{(iii) Content composition:} we limit the acceptable number of YOLO~\cite{yolo11_ultralytics} detections, discarding both trivially empty and pathologically cluttered scenes, and we drop images dominated by rendered text (OCR-derived text length and area share), collages, multi-panel layouts, and uniform synthetic backgrounds. \emph{(iv) Provenance artifacts:} a dedicated watermark detector and an EXIF scan remove stock-photography imagery. \emph{(v) VLM filtering:} a final VLM pass re-labels every surviving image for several categories, catching the missed cases. The full cascade retains $2.9$M images, i.e. $\sim\!22\%$ of the input. Figure~\ref{fig:filter_stats} shows the filtering pipeline in detail, along with distributions of several filtering statistics and example images drawn from the rejected regions of each distribution.

\subsubsection{Fake Image Generation}

\paragraph{Prompt synthesis.}
Each generated image originates from a prompt derived from a specific real image. This reduces the content gap between web photographs and images generated from independently written prompts, which could otherwise let detectors distinguish subject matter rather than synthesis artifacts.

We first obtain a dense image description with a VLM, then use an instruction-tuned LLM to rewrite it as a generation prompt. The instructions prioritize content type, subjects, object counts and relations, followed by scene, composition and lighting; preserve visible text; and prohibit invented details and quality adjectives. When shortening a prompt, the LLM drops lower-priority details first. We target $\sim35$--$50$ words so that most prompts fit within the $77$-token context of older CLIP-based text encoders. The full instructions are in Appendix~\ref{app:prompt-instructions}. Each prompt is assigned to $3$--$4$ randomly selected generators. We use the same prompt across generators, without model-specific rewriting, negative prompts or hyperparameter tuning, to avoid confounding generator identity with prompt style. The unfiltered pool contains $2.96$M prompts and $\sim8.4$M generated images.

\paragraph{Model selection.}
We select 33 open-source and 9 proprietary generators along three axes: generative paradigm, model scale and release recency, and inference-time compute regime. The set covers successive latent-diffusion models from the Stable Diffusion lineage; DiT-based PixArt models; cascaded pixel-space diffusion (DeepFloyd IF); flow-matching models from the FLUX family and other lineages; and autoregressive or unified multimodal generators. Including older and newer releases within the Stable Diffusion and Kandinsky lineages, alongside models from other research ecosystems and proprietary providers, broadens coverage across architectures and training pipelines. Appendix~\ref{app:dataset} and ~\ref{app:generator-inventory} provide the full model list and split assignments. We exclude older GAN-based generators: in preliminary experiments, DF-GAN and GALIP followed our image-derived prompts poorly and were detected nearly perfectly. Including them would reintroduce a content gap and inflate aggregate detection performance. 
We also include distilled and multi-step variants to examine how inference-time acceleration relates to detectability.
These comparisons let us measure detectability along the quality--latency frontier directly rather than infer it from proxy metrics.

\paragraph{Assignment of generators to splits.}
Generators are partitioned across splits by architectural family and by generation quality, so that detection difficulty increases monotonically along train~$\rightarrow$~val~$\rightarrow$~test~$\rightarrow$~Expert. The generator sets are deliberately \emph{not} disjoint: a controlled overlap between splits (e.g.\ FLUX.1-Kontext-dev, Infinity-8B, Ovis-Image and DeepFloyd~IF appear in all three base splits) makes it possible to separate seen-generator from unseen-generator performance for the same detector. Proprietary generators appear only in the Expert subset. Appendix~\ref{app:generator-inventory} lists all assignments.

\subsubsection{Dataset Sampling and Split Construction}

Before sampling, the merged real--generated pool of $10.60$M images passes
through a second filtering stage applied \emph{symmetrically} to both
classes. Any criterion that real and generated images satisfy at
systematically different rates is a potential shortcut. We therefore
derive thresholds on image characteristics from the \emph{real}
distribution and apply the same bounds to both classes. Generated
images additionally pass a VLM quality gate that removes generation
failures. This leaves $4.2$M images ($1.3$M real, $2.9$M generated). We use two sampling strategies: a density-based criterion for Expert,
which targets images that are statistically hardest to attribute, and
stratified cluster-based sampling for Base, which preserves the
diversity of the source distribution.

\paragraph{Feature representation.}
Both criteria operate in a shared space combining semantic content,
object detections, content-type tags, and image quality and structural
features. Directly estimating densities in this heterogeneous space
would let high-dimensional blocks such as CLIP dominate distances.
We therefore reduce and rescale the feature blocks before projecting
them to $12$ dimensions. The projection is fit on a class-balanced
subsample; otherwise, the larger generated class would tilt the
principal axes towards its distribution. Details are in Appendix~\ref{app:sampling}.

\begin{figure*}[tbp]
    \centerline{\includegraphics[width=0.99\textwidth]{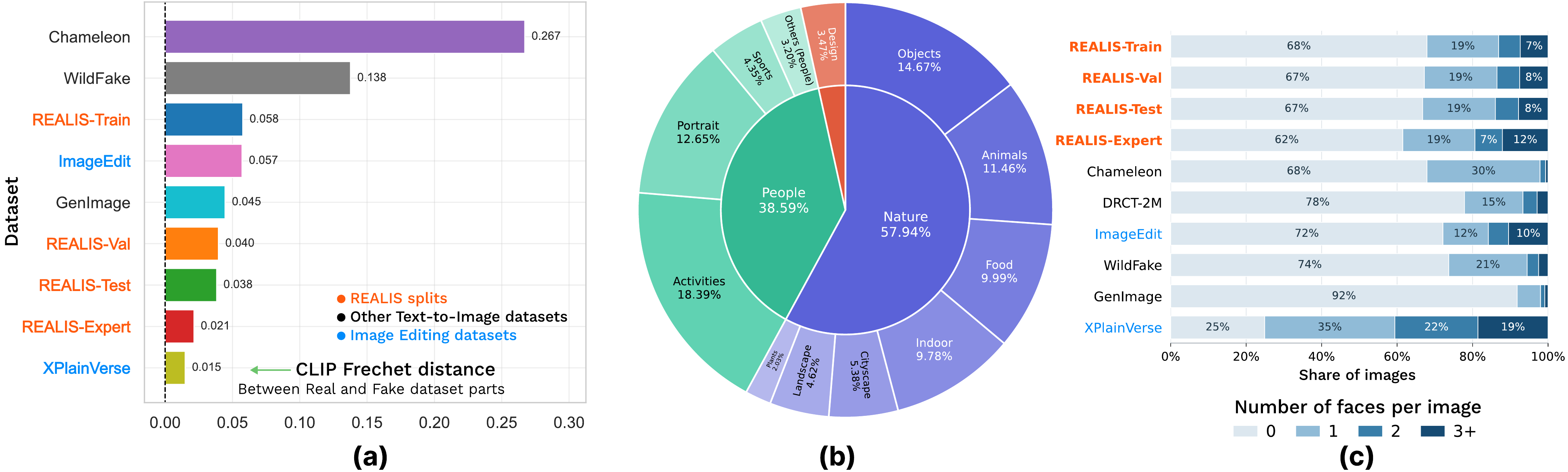}}
    \caption{Class alignment and content composition of REALIS splits and other datasets. \textbf{(a)} Fr\'echet distance between real and generated CLIP-embedding distributions; lower values indicate closer class distributions. \textbf{(b)} Hierarchical content taxonomy of the pooled REALIS evaluation splits, assigned by a VLM; inner and outer sectors show broad categories and their subdivisions. \textbf{(c)} Proportions of images containing different numbers of faces, as detected with the MTCNN~\cite{zhang2016mtcnn} model. REALIS combines closely aligned real and generated distributions with varied semantic and scene composition.}
\label{fig:dataset_stats}
\vspace{-10pt}
\end{figure*}

\paragraph{REALIS-Expert.}
We select hard images using the local log density ratio
$\log(p_{\mathrm{real}}(x)/p_{\mathrm{gen}}(x))$. Images with ratios near
zero lie in regions where the two classes occur at similar density in the
chosen feature space, removing much of the semantically and photometrically
easy portion of the distribution. This guarantee is feature-space specific:
low-level synthesis fingerprints may still separate the classes in pixel space.
The criterion is therefore generator- and detector-agnostic.

We build separate HNSW indices over the 12-D projections of the real and
generated pools and estimate both densities with the
Loftsgaarden--Quesenberry $k$-NN estimator. Retaining images with
$|\log\hat r(x)|\leq0.5$ keeps $\sim\!12\%$ of the pool
($\sim\!500$k images) and lowers mean ROC-AUC across pretrained detectors
from $0.826$ to $0.809$. Restricting to the hardest generators lowers it
to $0.723$, and requiring a VLM quality score of at least $8$ on every
axis yields $0.716$, compared with $0.871$ on the training split.
REALIS-Expert is sampled uniformly from this pool, giving $130$k images
($60$k real, $70$k generated) across $16$ open-source and proprietary
generators.

We deliberately do \emph{not} use detector predictions to select Expert
images, since this would encode the blind spots of current detectors rather
than properties of the data. As a sanity check, $|\log\hat r(x)|$ is
negatively correlated with per-image detection error on generated images
across the tested architectures, with PLCC reaching $-0.15$.

\paragraph{Base splits.}
For train, validation, and test, the objective is the opposite: to
preserve diversity across the source distribution. We partition the
same $12$-D representation into $20$ $K$-means clusters, with
$k=20$ selected by standard cluster-quality criteria, and sample with
equal quotas per cluster. We use $55$k images per cluster for training
and $2.5$k real plus $2.5$k generated images per cluster for validation
and test, preventing popular content types from crowding out sparser
regions. Sampling respects each split's assigned generators.
REALIS-Base contains $1.1$M training images and balanced $100$k
validation and $100$k test sets over $30$ open-source generators.

\paragraph{Split disjointness.}
Splits are made disjoint at the level of \emph{prompts}, not images.
Each real image induces one prompt and $3$--$4$ near-isosemantic
generated counterparts; splitting them across train and test would
inflate reported generalization. A real image and all its generated
counterparts therefore always belong to exactly one split. 

\subsubsection{Image Degradation Pipeline}

Images encountered in the wild are almost never bit-exact generator output: they
have been recompressed, resized, color-graded, watermarked, screenshotted or
deliberately perturbed. To evaluate detectors under these conditions, we build a
degradation pipeline and apply it to the validation, test and Expert splits,
producing a distorted variant of each; detection performance is reported on both
the clean and the distorted version of every split.

The pipeline covers 35 distinct transformations organized into eleven groups: blur, color distortion, algorithmic compression, neural compression, multiple and mixed recompression chains, noise, brightness changes, geometric transforms, spatial distortions, invisible watermarks and some adversarial attacks. The Appendix \ref{app:transformations} provides a detailed list of distortions. Each transformation is parameterized by five
strength levels, calibrated so that level~1 is barely perceptible, while level~5
is severe but still leaves the image recognizable.

Rather than applying a single transformation, we simulate realistic processing
histories by chaining several. For each image we draw $1$--$5$ groups without
replacement, pick one transformation from each and apply them in random order,
with the strength level drawn from a discretized Gaussian over the five levels
so that moderate degradations dominate while both extremes remain rare. The exact chain and the strength of each element are recorded per
image, so robustness can be analyzed per transformation, per group, and per
strength level in addition to being reported in aggregate.

For detectors trained on the REALIS-Train split, we employ a restricted augmentation set
--- essentially the classical group (JPEG, blur, noise, crop, color) --- while
the neural codecs, recompression chains, adversarial attacks and watermarking methods are held out entirely. A detector trained on our
training split therefore still faces genuinely unseen degradations at evaluation
time, which is what makes the robustness numbers informative rather than a
measure of augmentation coverage.

\subsection{Benchmark}
We evaluate three groups of detectors: off-the-shelf specialized AI-image detectors, the same detector family fine-tuned on REALIS-Train, and zero-shot vision-language models. We additionally evaluate non-tuned detectors on GenImage, Chameleon, and WildFake to measure how rankings transfer from existing benchmarks to REALIS. Our primary metric is ROC-AUC, which avoids committing to detector-specific score thresholds. For analyses spanning multiple generators, we report generator-macro ROC-AUC by computing AUC separately for each generator against a shared real-image pool and averaging uniformly across generators. We additionally report false-positive and false-negative rates, as well as recall and F1 metrics, for analyses of operating-point behavior.

\section{Results}

\subsection{Dataset Content, Diversity, and Complexity}
\label{sec:dataset_content_diversity}

REALIS combines broad semantic and structural diversity with increasingly aligned real and synthetic distributions. Its evaluation splits cover diverse content types, object and face counts, scene complexity, and aspect ratios (Figure~\ref{fig:dataset_stats}c), while all splits occupy a broad region of the pooled CLIP embedding space (Figure~\ref{fig:clip_tsne}). Vendi scores and Participation Ratios further show that all REALIS splits exceed the compared T2I benchmarks on both diversity measures, with REALIS-Expert achieving the highest Vendi score. At the same time, the Fr\'echet distance between real and generated CLIP embeddings decreases from train to Expert (Figure~\ref{fig:dataset_stats}b). REALIS-Expert has the smallest gap among the compared text-to-image detection datasets, second only to XPlainVerse, whose generated images are produced by editing real images. Thus, REALIS-Expert preserves broad content coverage while reducing coarse real--synthetic distribution differences, supporting evaluation beyond simple semantic shortcuts. A small Fr\'echet distance alone, however, does not eliminate residual shortcuts.

\begin{figure*}[t]
    \centering
    \includegraphics[width=\textwidth]{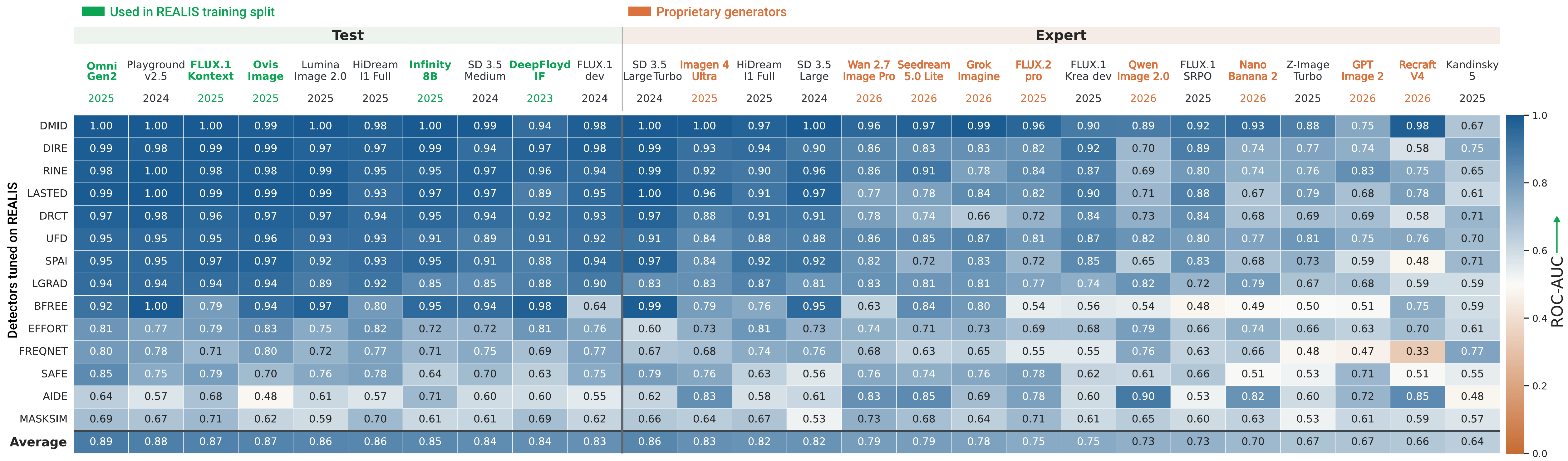}
    \caption{
    Per-generator ROC-AUC of 14 REALIS-tuned detectors against a shared real-image pool from Test and Expert. Green/orange labels mark training/proprietary generators; release years appear below. Generators are sorted by mean AUC within each split; the bottom row averages detectors.
    }
    \label{fig:per_gen_results_tests}
\vspace{-10pt}
\end{figure*}

\subsection{Overall Performance}

Table~\ref{tab:non_tuned_detectors} compares off-the-shelf detectors and their
REALIS-tuned variants on three external benchmarks and six REALIS evaluation
settings. Table~\ref{tab:all_distorted_results} compared results for off-the-shelf VLM models on REALIS-Expert. Off-the-shelf results on clean Test are unavailable and are marked
with dashes. Results for Expert and Expert-distorted cover the open-source
component of REALIS-Expert.
Strong performance on external benchmarks does not necessarily transfer:
off-the-shelf RINE achieves 97.34\% ROC-AUC on GenImage but 44.3\% on
Expert-distorted. REALIS-tuned DMID reaches 98.5\% ROC-AUC on clean Test,
while Expert-distorted remains challenging. The REALIS-tuned ProDet run
failed to converge, and its near-chance REALIS AUCs should not be interpreted
as successful fine-tuning.

\subsection{Generalization Across Generators}
\label{sec:generator_generalization}

\textbf{Transfer is strongly generator-dependent.} Figure~\ref{fig:per_gen_results_tests} compares REALIS-tuned detectors using a real-image reference pooled from Test and Expert. Many approaches saturate on Test, but Expert results vary widely: SD~3.5 Large-Turbo and Imagen~4 Ultra remain relatively easy, while Kandinsky~5 and Z-Image-Turbo challenge most detectors. Training exposure does not explain this ordering; held-out Playground~v2.5 is among the easiest Test generators, with higher AUC than several generators represented in training. REALIS thus reveals both transfer to unseen generators and failures obscured by aggregate scores.

\textbf{Shared components do not imply shared detectability.} Kandinsky~5 and Z-Image-Turbo share the FLUX autoencoder but use cross-attention and single-stream diffusion transformers, respectively~\cite{arkhipkin2025kandinsky,cai2025z}. Their lower average AUCs than FLUX.1-dev suggest that autoencoder sharing alone does not predict transfer. Acceleration is likewise not a universal indicator of difficulty: distilled SD~3.5 Large-Turbo is easier to detect than its Large counterpart, while Z-Image-Turbo remains challenging. These comparisons motivate evaluation of individual synthesis pipelines but cannot isolate the effects of architecture, training setup, or sample selection.

\textbf{Detector rankings and operating thresholds also matter.} FREQNET is weak on many generators but achieves the highest plotted AUC on Kandinsky~5. High AUC need not imply high recall at a fixed threshold: on FLUX.1-dev SRPO, DMID reaches about $0.92$ AUC but detects only $35\%$ of images at the default $0.5$ threshold (Appendix~\ref{app:fixed-threshold}, Figure~\ref{fig:recall_per_gen}). Operating-point reliability therefore requires interpreting recall alongside false-positive rates.

\begin{table*}[t!]
\centering
\caption{Comparison of off-the-shelf (OFT) and REALIS-tuned (FT) detector performance across datasets. All values are ROC-AUC in percent. 
}
\label{tab:non_tuned_detectors}
\resizebox{\textwidth}{!}{%
\begin{tabular}{l*{18}{r}}
\toprule
\textbf{Detector}
 & \multicolumn{2}{c}{\textbf{GenImage}}
 & \multicolumn{2}{c}{\textbf{Chameleon}}
 & \multicolumn{2}{c}{\textbf{WildFake}}
 & \multicolumn{12}{c}{\textbf{REALIS}} \\
\cmidrule(lr){8-19}
 & & & & & &
 & \multicolumn{2}{c}{Val}
 & \multicolumn{2}{c}{Val-distorted}
 & \multicolumn{2}{c}{Test}
 & \multicolumn{2}{c}{Test-distorted}
 & \multicolumn{2}{c}{Expert}
 & \multicolumn{2}{c}{Expert-distorted} \\
\cmidrule(lr){2-3}\cmidrule(lr){4-5}\cmidrule(lr){6-7}
\cmidrule(lr){8-9}\cmidrule(lr){10-11}\cmidrule(lr){12-13}
\cmidrule(lr){14-15}\cmidrule(lr){16-17}\cmidrule(lr){18-19}
 & \multicolumn{1}{c}{OTS} & \multicolumn{1}{c}{FT}
 & \multicolumn{1}{c}{OTS} & \multicolumn{1}{c}{FT}
 & \multicolumn{1}{c}{OTS} & \multicolumn{1}{c}{FT}
 & \multicolumn{1}{c}{OTS} & \multicolumn{1}{c}{FT}
 & \multicolumn{1}{c}{OTS} & \multicolumn{1}{c}{FT}
 & \multicolumn{1}{c}{OTS} & \multicolumn{1}{c}{FT}
 & \multicolumn{1}{c}{OTS} & \multicolumn{1}{c}{FT}
 & \multicolumn{1}{c}{OTS} & \multicolumn{1}{c}{FT}
 & \multicolumn{1}{c}{OTS} & \multicolumn{1}{c}{FT} \\
\midrule
AIDE
 & 8.12 & 73.12 & 47.99 & 48.80 & 15.83 & 74.43
 & 49.9 & 64.1 & 50.5 & 65.5 & \na & 60.1
 & 50.9 & 64.0 & 50.1 & 57.8 & 51.7 & 58.6 \\
DIRE
 & 41.21 & 82.71 & 54.96 & 68.63 & 65.51 & 40.67
 & 39.5 & 98.4 & 40.6 & 80.1 & \na & 97.3
 & 41.4 & 77.9 & 47.9 & 86.9 & 48.7 & 65.7 \\
DRCT
 & 67.04 & 80.29 & 68.64 & 71.01 & \na & 64.55
 & 74.1 & 96.3 & 59.3 & 80.8 & \na & 94.9
 & 57.5 & 78.4 & 65.6 & 82.6 & 53.5 & 68.5 \\
EFFORT
 & 56.06 & 62.68 & 65.87 & 58.44 & \textbf{89.86} & 48.64
 & 67.7 & 80.4 & 59.0 & 75.9 & \na & 79.1
 & 56.3 & 74.3 & 55.3 & 66.7 & 53.1 & 63.7 \\
FreqNet
 & 64.69 & 66.93 & 31.06 & 57.31 & 89.04 & 47.46
 & 49.6 & 75.9 & 51.7 & 58.7 & \na & 75.8
 & 51.5 & 59.0 & 41.8 & 63.3 & 50.3 & 52.7 \\
MaskSim
 & 61.94 & 67.28 & 61.87 & 64.95 & 46.07 & 50.98
 & 62.1 & 62.9 & 52.7 & 59.6 & \na & 65.5
 & 52.7 & 61.7 & 53.4 & 58.0 & 51.2 & 55.3 \\
ProDet
 & 56.43 & 58.70 & 47.62 & 39.50 & 53.42 & 0.00
 & 58.2 & 48.3 & 54.6 & 47.6 & \na & 48.3
 & 54.5 & 47.8 & 53.9 & 50.4 & 52.0 & 49.9 \\
RINE
 & \textbf{97.34} & 88.16 & 51.40 & 54.35 & 87.44 & 79.50
 & 53.6 & 97.5 & 45.7 & 75.1 & \na & 96.3
 & 45.3 & 72.2 & 54.7 & 85.1 & 44.3 & 63.4 \\
SAFE
 & 89.12 & 78.05 & 54.63 & 72.17 & 86.29 & 49.78
 & 44.6 & 68.9 & 46.3 & 57.5 & \na & 69.4
 & 46.4 & 57.7 & 46.9 & 62.8 & 48.9 & 55.5 \\
SPAI
 & 85.39 & 84.70 & 57.94 & 64.26 & 73.0 & 30.11
 & 79.4 & 94.8 & 58.0 & 70.3 & \na & 93.7
 & 57.1 & 68.8 & 73.7 & 82.7 & 54.1 & 59.4 \\
UFD
 & 80.09 & 71.22 & 43.25 & \textbf{74.22} & 80.83 & 55.01
 & 54.6 & 94.0 & 49.1 & \textbf{87.6} & \na & 92.8
 & 47.6 & \textbf{86.3} & 44.0 & 82.3 & 42.2 & \textbf{75.2} \\
\bottomrule
\end{tabular}%
}
\vspace{-10pt}
\end{table*}

\subsection{Performance Across Content and Image Characteristics}
\label{sec:content_statistics_performance}

\begin{figure*}[t]
    \centering
    \includegraphics[width=\textwidth]{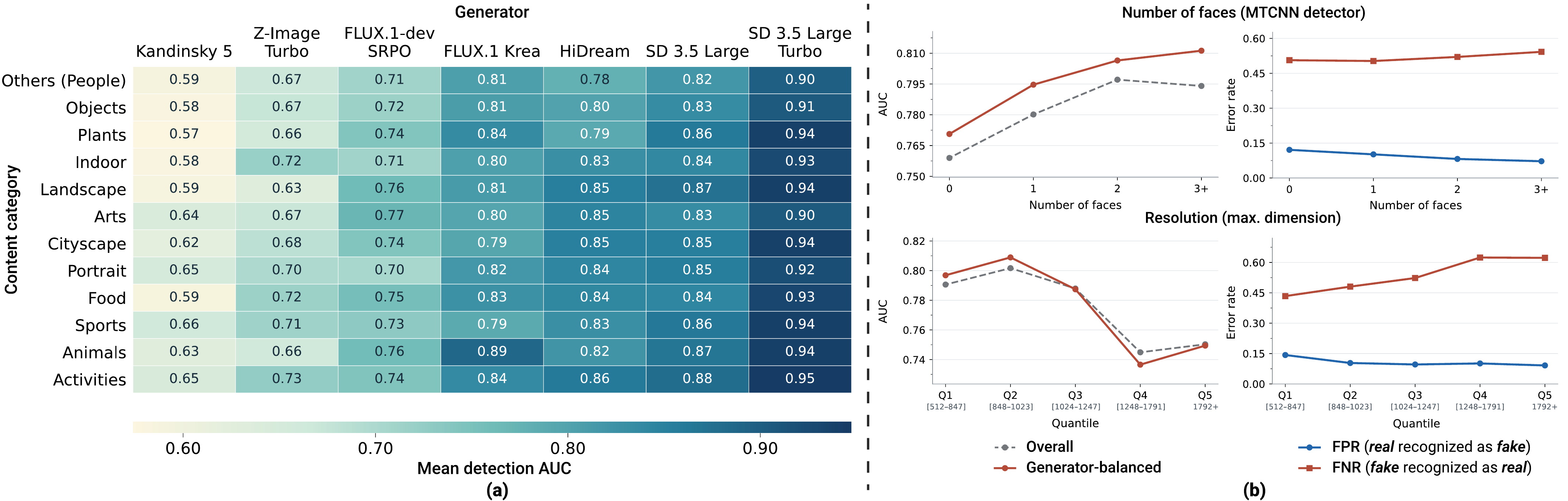}
    \caption{Relationship between image content and characteristics and average detector efficiency (across 16 best-performing models), evaluated on REALIS-Expert split. Proprietary generators are excluded due to lower number of samples per model. (a) Detection quality across various VLM-classified content types for different generators. (b) ROC-AUC, FPR and FNR variation across different image characteristics. Resolution is binned into 5 groups using quantiles (20th, 40th, etc.). For ROC-AUC, both per-generator-balanced and overall generator-agnostic values are reported.}
    \label{fig:content_analysis}
    \vspace{-10pt}
\end{figure*}

\textbf{Generator differences exceed content variations.}
Figure~\ref{fig:content_analysis} analyzes the seven open-source Expert generators, averaging over 16 reference detectors spanning tuned methods, off-the-shelf methods, and zero-shot VLMs. Both real and generated images are restricted to the same content category or statistic bin; generator-balanced AUC then weights eligible generators equally. In Figure~\ref{fig:content_analysis}(a), the range of category-averaged AUC across generators is approximately $0.31$, compared with $0.05$ for generator-averaged AUC across content categories. Kandinsky~5 remains difficult, and SD~3.5 Large-Turbo is comparatively easy throughout the taxonomy. Thus, generator identity is the stronger correlate of detection difficulty in this analysis, although content still introduces smaller yet measurable variation.

\textbf{Generator composition explains many apparent statistic effects.}
The extended sweeps in the Appendix \ref{app:balanced-sweeps} show that associations with estimated noise variance, high-frequency energy, and aesthetic scores shrink markedly when generator contributions are balanced. Consequently, pooled trends can largely reflect which generators populate each bin. Other associations persist: images with more detected objects have higher generator-balanced AUC, and the face-count analysis in Figure~\ref{fig:content_analysis}(b) also shows a modest increase. At the default threshold of $0.5$, however, additional faces accompany fewer false positives on real images, while the false-negative rate stays similar or increases. Better AUC therefore does not imply that generated images containing more faces are more often flagged.

\textbf{Resolution exposes a different error pattern.}
In Figure~\ref{fig:content_analysis}(b), the two highest resolution bins have lower AUC and substantially more missed synthetic images, despite relatively low false-positive rates. This association remains visible after generator balancing, yet it can also be affected by intrinsic relationships between image resolution and depicted content, limiting its interpretation to the evaluated distributions.  
Together, these results show why content and image statistics remain useful diagnostic axes even when generator differences dominate: they reveal distinct false-alarm and missed-detection patterns that a single aggregate score cannot describe.

\label{sec:postprocessing_robustness}

\subsection{Robustness to Post-Processing}
\label{sec:postprocessing_robustness}

\begin{figure*}[t]
    \centering
    \includegraphics[width=\textwidth]{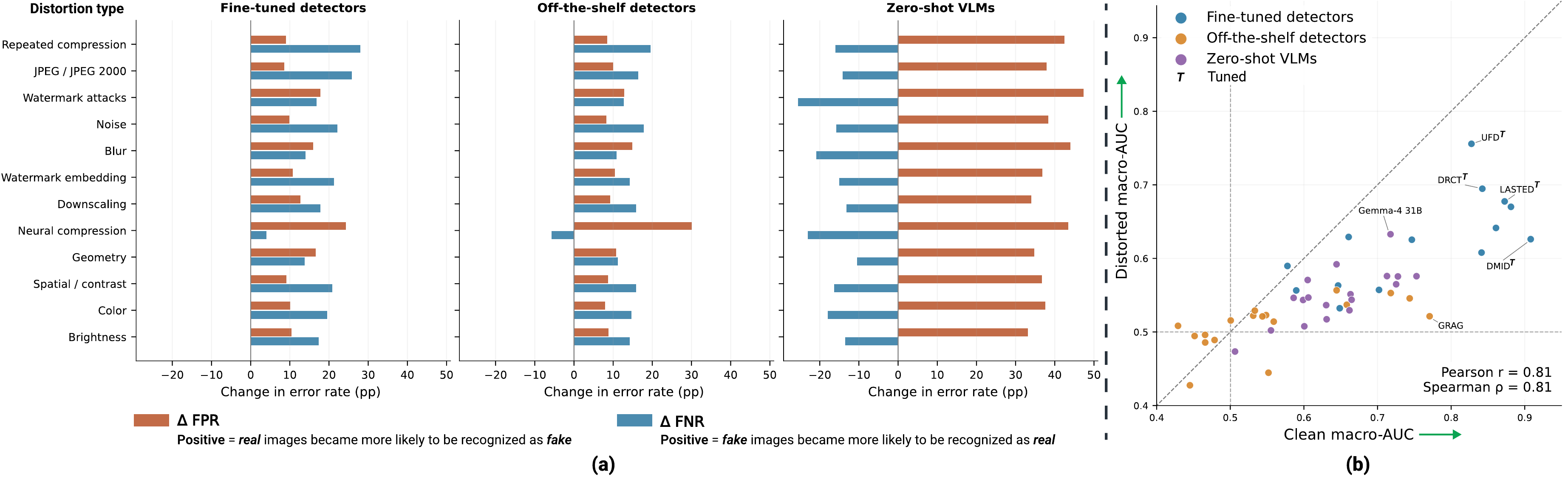}
    \caption{
    Post-processing robustness on REALIS-Expert. \textbf{(a)} Changes in FPR and FNR at threshold 0.5 relative to matched clean images, averaged over 14 representative detectors for subsets containing each distortion group. Positive values indicate more errors; transformations may co-occur. \textbf{(b)} Clean versus distorted generator-macro ROC-AUC for all evaluated detectors, using paired predictions.
    }
    \label{fig:postprocessing_overview}
    \vspace{-4pt}
\end{figure*}

\textbf{Clean performance predicts only part of post-processing robustness.}
We compare clean images with their post-processed counterparts in the open-source component of REALIS-Expert. Table~\ref{tab:non_tuned_detectors} shows sharp detection losses even for detectors with strong clean-image performance, with many approaching random guessing. Using generator-balanced AUC, clean and processed performance correlate across all detectors (Spearman $\rho=0.81$; Figure~\ref{fig:postprocessing_overview}b), but rankings can shift: REALIS-tuned UFD trails DMID on clean images yet retains about $0.76$ AUC after processing versus DMID's $0.63$. Near-chance detectors performance may change slightly without becoming reliable. Clean performance and retained performance therefore offer complementary measures of robustness.

\textbf{Post-processing induces different error biases across detector groups.}
Figure~\ref{fig:postprocessing_overview}a shows error changes at the default $0.5$ threshold for 14 reference detectors selected by clean-image performance. Conventional detectors generally incur more false positives and false negatives; repeated compression and JPEG degradation especially increase misses. Across every distortion group, the VLM instead produces more false positives and fewer false negatives, indicating a shift toward the generated label despite poorer class separation. Neural compression shows a similar asymmetry among conventional detectors: false-positive increases dominate, while off-the-shelf detectors also miss fewer generated images. Recall or aggregate accuracy alone would obscure these patterns.

\textbf{Transformation type and strength expose complementary vulnerabilities.} Appendix~\ref{app:postprocessing_analysis} identifies mixed recompression, glass blur, and shot noise among the largest AUC losses. Increasing distortion severity generally worsens discrimination, especially for blur and JPEG compression, while repeated recompression is damaging even at its lowest setting. A separate validation-sample ablation shows a large drop when the distortion pipeline is enabled and further losses as its strength increases. Expert-split comparisons use clean versions of the same images; because transformations can co-occur, they characterize complete processing chains rather than individual stages' causal effects.

\begin{table*}[t]
\vspace{-10pt}
\centering
\caption{Zero-shot VLM ROC-AUC on REALIS-Expert and its distorted variant.
Median AUC across models is 0.625 on Expert and 0.538 on Expert-Distorted.}
\label{tab:all_distorted_results}
\scriptsize
\setlength{\tabcolsep}{2pt}
\renewcommand{\arraystretch}{0.99}
\begin{tabular}{lcc@{\hspace{1em}}lcc@{\hspace{1em}}lcc}
\toprule
Model & Expert & Expert-Dist. &
Model & Expert & Expert-Dist. &
Model & Expert & Expert-Dist. \\
\midrule

Qwen3-VL 235B-A22B
& \textbf{0.734} & 0.562
& Qwen2.5-VL 32B
& 0.645 & 0.539
& InternVL3.5 30B-A3B
& 0.597 & 0.543 \\

Qwen3-VL 32B
& 0.712 & 0.566
& Qwen2.5-VL 7B
& 0.617 & 0.508
& InternVL3.5 14B
& 0.590 & 0.537 \\

Qwen3-VL 30B-A3B
& 0.688 & 0.560
& Qwen2.5-VL 3B
& 0.584 & 0.499
& InternVL3.5 8B
& 0.536 & 0.490 \\

Qwen3-VL 30B-A3B (think)
& 0.636 & 0.533
& Gemma-4 31B
& 0.705 & \textbf{0.622}
& InternVL3.5 4B
& 0.492 & 0.464 \\

Qwen3-VL 8B
& 0.706 & 0.555
& Gemma-4 12B
& 0.633 & 0.584
& & & \\

Qwen3-VL 8B (think)
& 0.613 & 0.522
& Gemma-4 4B
& 0.570 & 0.536
& & & \\

Qwen3-VL 4B
& 0.639 & 0.516
& Gemma-4 2B
& 0.594 & 0.565
& & & \\

\bottomrule
\end{tabular}
\vspace{-10pt}
\end{table*}

\vspace{-4pt}
\section{Conclusion}
We introduce REALIS, a dataset of 1.43 million images spanning 42 text-to-image generators, designed to study detection under generator and post-processing shifts. Its construction reduces semantic and quality differences between real and synthetic images while preserving content diversity. REALIS-Expert further targets closely matched class distributions without using detector predictions for selection, and paired clean and processed images support controlled robustness evaluation. Our experiments show that strong performance on existing benchmarks or clean images does not ensure reliable detection after processing, with substantial degradation affecting both specialized detectors and zero-shot VLMs. Fine-tuning on REALIS improves performance, yet the best tuned detector reaches only 0.752 ROC-AUC on the processed open-source component of REALIS-Expert. Processing also changes error patterns, including increased false positives on real images, highlighting the need to assess false alarms alongside missed detections. Together, these findings motivate joint evaluation of generator generalization and processing robustness, with REALIS providing a common resource for developing and diagnosing detectors under both challenges.

\subsection*{AI use statement}

In this work, generative AI is part of the methodology. The synthetic part of REALIS was produced by the text-to-image models listed in Appendix C; dense VLM captions of real images were rewritten into generation prompts with Qwen3-30B-A3B (Appendix B); VLMs, including InternVL3.5-38B, were used to filter, score and tag images during dataset construction (Section 3.1); and VLMs are evaluated as zero-shot detectors (Section 3.2). We additionally used generative AI tools to polish the writing and to assist literature search. All AI-assisted outputs, including references, were checked manually, and the authors take full responsibility for the content of this work.

\subsection*{Reproducibility statement}

A subset of REALIS is available for peer review at
\url{https://anonymous-hf.com/a/89q3tdboxa2m/}.
Due to the practical challenges of fully anonymizing the dataset for
double-blind review, we provide only a partial release during the review
period. The full dataset will be released upon acceptance.
The dataset construction procedure is described in
Section~\ref{sec:dataset_construction} and further detailed in the appendices.
Code for dataset generation and detector evaluation is available at
\url{https://anonymous-hf.com/a/jshtpiysv3ua/}.

\bibliography{iclr2027_conference}
\bibliographystyle{iclr2027_conference}

\appendix

\section{Contents}
Appendix~\ref{app:dataset} details relevant dataset comparison~\ref{app:comparison}, prompt synthesis~\ref{app:prompt-synthesis}, generator selection~\ref{app:generators}, data filtering, and split construction~\ref{app:sampling}. 

Appendix~\ref{app:prompt-instructions} gives the prompt-generation instructions.

Appendix~\ref{app:generator-inventory} lists generators by split.

Appendix~\ref{app:transformations} specifies the post-processing pipeline. 

Appendix~\ref{app:detector-results} extends the detector results with per-generator~\ref{app:per-generator-results} and fixed-threshold evaluations~\ref{app:fixed-threshold}. 

Appendix~\ref{app:image-characteristics} examines image characteristics~\ref{app:balanced-sweeps}, generation quality, and generator profiles~\ref{app:generator-quality}. 

Appendix~\ref{app:postprocessing_analysis} analyzes sensitivity to post-processing transformations and their strength, including a separate validation-sample ablation.

\section{Dataset details}
\label{app:dataset}

\begin{figure*}[tbp]
    \centerline{\includegraphics[width=0.99\textwidth]{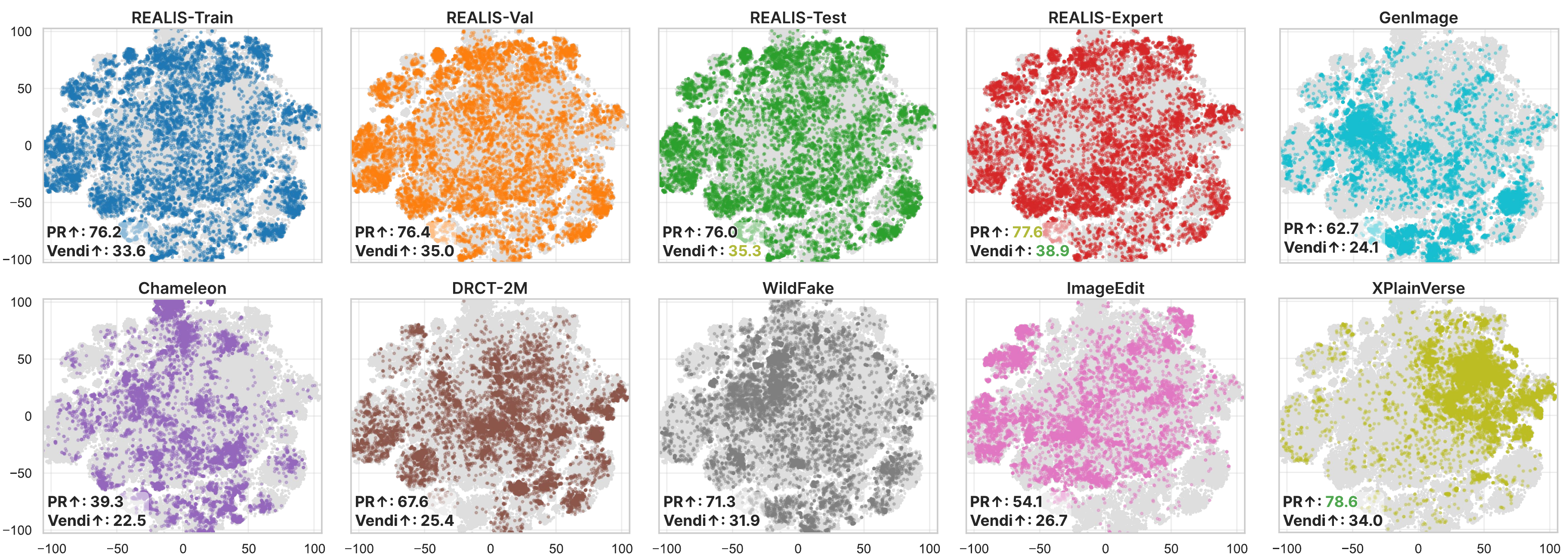}}
    \caption{2D t-SNE representation of CLIP-space coverage for different deepfake detection datasets, alongside their Vendi Scores~($\uparrow$)~\cite{vendi_metric} and Participation Ratios~($\uparrow$)~\cite{pr_metric} to quantify the diversity. Scores are calculated over original CLIP embeddings on full datasets; t-SNE projection is used only for visualization. Grey dots represent the concatenation of all datasets, with random subsamples of similar size~(10k) used for each entry.}
\label{fig:clip_tsne}
\end{figure*}

\begin{table*}[t]
\centering
\tiny
\setlength{\tabcolsep}{2.8pt}
\renewcommand{\arraystretch}{1.10}
\caption{
Comparison with representative AI-generated image detection datasets and benchmarks.
\#Gen. denotes the number of fake-image generators or generation sources when reported.
\#Closed denotes proprietary/API/community-service generators whose weights are not publicly released.
``---'' indicates that the paper or dataset page does not report a clean value. Robust. stands for robustness to processing techniques.
}
\label{tab:dataset_comparison}
\resizebox{\textwidth}{!}{%
\begin{tabular}{
@{}
p{3.05cm}
r
r
r
r
r
p{1.45cm}
p{1.00cm}
p{2.85cm}
p{1.45cm}
@{}
}
\toprule
\textbf{Dataset / Benchmark} &
\textbf{Year} &
\textbf{Real} &
\textbf{Fake} &
\textbf{\#Gen.} &
\textbf{\#Closed} &
\textbf{Scope} &
\textbf{Access} &
\textbf{Latest Generator} &
\textbf{Robust.} \\
\midrule

ForenSynths~\cite{wang2020cnnspot}
& 2020 & 362K & 362K & 11 & 0 & GAN-era & \bad{Hard} & StyleGAN2 (2019) & \midc{Basic} \\

Diff.Forensics~\cite{wang2023dire}
& 2023 & 140K & 570K & 8 & 0 & Scene & \bad{Hard} & SD v2 (2022) & \bad{No} \\

DMDetection~\cite{corvi2023dm}
& 2023 & 200K & 200K & 3 & 1 & Scene & \bad{Hard} & DALL-E 2 (2022) & \midc{Basic} \\

DE-FAKE~\cite{sha2023defake}
& 2023 & \na & \na & 4 & 1 & T2I & \bad{Hard} & DALL-E 2 (2022) & \bad{No} \\

GenImage~\cite{zhu2023genimage}
& 2023 & 1330K & 1350K & 8 & 1 & Object & \bad{Hard} & Midjourney (2022) & \midc{Basic} \\

ArtiFact~\cite{rahman2023artifact}
& 2023 & 965K & 1530K & 25 & 0 & Multi-cat. & \good{Easy} & SD v2 (2022) & \midc{Basic} \\

Synthbuster~\cite{bammey2023synthbuster}
& 2023 & 1K & 9K & 9 & 4 & Prompt & \good{Easy} & DALL-E 3 (2023) & \bad{No} \\

CiFAKE~\cite{bird2023cifake}
& 2023 & 60K & 60K & 1 & 0 & CIFAR-10 & \good{Easy} & LD (2022) & \bad{No} \\

Fake2M~\cite{lu2023fake2m}
& 2023 & 1090K & 2080K & 3 & 0 & Perception & \bad{Hard} & IF (2023) & \midc{Basic} \\

SIDBench~\cite{zhong2023aigcdetect}
& 2024 & 406K & 412K & 17 & 2 & General & \good{Easy} & DALL-E 2 (2022) & \midc{Basic} \\

DRCT-2M~\cite{chen2024drct}
& 2024 & \na & 2000K & 16 & 0 & Diffusion & \good{Easy} & SDXL (2023) & \midc{Basic} \\

Semi-Truths~\cite{pal2024semitruths}
& 2024 & 26K & 1340K & 8 & \na & General & \good{Easy} & SDXL (2023) & \bad{No} \\

DiffusionFace~\cite{chen2024diffusionface}
& 2024 & 30K & 600K & 11 & 0 & Face & \bad{Hard} & SD v2.1 (2022) & \bad{No} \\

DiFF~\cite{cheng2024diff}
& 2024 & 23K & 500K & 13 & 1 & Face & \bad{Hard} & Midjourney 5 (2023) & \bad{No} \\

WildFake~\cite{hong2025wildfake}
& 2025 & 1010K & 2560K & 23 & \na & Wild & \bad{Hard} & Community (---) & \midc{Basic} \\

Chameleon~\cite{liu2025chameleon}
& 2025 & 15K & 11K & \na & \na & Wild & \midc{Gated} & DALL-E 3 (2023) & \good{Hard set} \\

AI-GenBench~\cite{pellegrini2025aigenbench}
& 2025 & 180K & 180K & 36 & 3 & General & \good{Easy} & FLUX.1 (2024) & \good{Temporal} \\

TrueFake~\cite{dellanna2025truefake}
& 2025 & 160K & 440K & 8 & 2 & Social & \good{Easy} & FLUX.1 (2024) & \good{Sharing} \\

C.Forensics~\cite{park2025community}
& 2025 & 2700K & 2700K & 4803 & 11 & Gen.-rich & \good{Easy} & Imagen 3 (2024) & \good{Gen. stress} \\

RRDataset~\cite{li2025rrbench}
& 2025 & 10K & 10K & \na & \na & Scenario & \good{Easy} & FLUX.1 (2024) & \good{Re-digit.} \\

OpenFake~\cite{livernoche2025openfake}
& 2025 & 3000K & 963K & 18 & 7 & Political & \good{Easy} & Imagen 4 (2025) & \good{Arena} \\

MS COCOAI~\cite{roy2026mscocoai}
& 2026 & 16K & 80K & 5 & 2 & COCO & \good{Easy} & Midj. v6 (2023) & \midc{Basic} \\

\midrule
\textbf{Ours}
& \ourscell{2026}
& \ourscell{2200K}
& \ourscell{8800K}
& \ourscell{42}
& \ourscell{12}
& \ourscell{General}
& \ourscell{Easy}
& \ourscell{\mbox{GPT Image 2 (2026)}}
& \ourscell{Extensive} \\

\bottomrule
\end{tabular}%
}
\end{table*}

\subsection{Comparison with Existing Datasets}
\label{app:comparison}

Table~\ref{tab:dataset_comparison} summarizes representative AI-generated image detection datasets by size, generator coverage, content scope, access, and robustness evaluation. These attributes describe different design choices rather than a single ranking: for example, some benchmarks prioritize the number of generation sources, while others focus on temporal generalization or transformations encountered during image sharing. Our dataset combines 2.2 million real and 8.8 million synthetic images from 42 generation sources, including 12 closed models, with general-content coverage and extensive post-processing evaluation. Reported counts and qualitative categories follow the respective dataset descriptions and may reflect different counting conventions.

\subsection{Prompt synthesis}
\label{app:prompt-synthesis}
Every generated image in REALIS originates from a prompt that describes a
specific real image from the sourced pool. This design eliminates the content
gap that otherwise separates the two classes in detection datasets, where real
images are web photographs while generated images follow the very different
distribution of user-written prompts. Under such a gap, a detector can reach high
accuracy by recognizing \emph{subject matter} rather than \emph{synthesis
artifacts}, and its reported accuracy says little about what it will do in
deployment.

Prompts are produced in two steps. First, a dense description of the real image
is obtained with a VLM. Second, an
instruction-tuned LLM rewrites
that description into a generation prompt. The LLM is instructed to follow
standard prompt-construction practice--- lead with content type, exact object
counts, subjects and their main relation, then add scene, composition and
lighting; reproduce any on-image text verbatim together with its location;
invent no details and no quality adjectives --- and to drop material in a fixed
tail-priority order when the length budget is exceeded, never dropping content
type, entities, counts, or relations. We target a compact prompt of
$\sim\!35-50$ words: long enough to pin down the scene, short enough that
most of the prompts fit inside the $77$-token context of CLIP-based text
encoders and are therefore rendered without truncation even by the older models
in our pool. The full instruction template is given in
Appendix~\ref{app:prompt-instructions}.

Each prompt is then assigned $3$ to $4$ randomly chosen generators, so that in
the unfiltered pool every real image has $3$--$4$ content-matched synthetic
counterparts. We deliberately issue the same prompt to every generator, with no
model-specific rewriting, negative prompts or per-model hyperparameter tuning:
such tuning would confound generator identity with prompt style and make
cross-generator comparison unreliable. The full pool comprises $2.96$M prompts
and $\sim\!8.4$M generated images.

\subsection{T2I models selection}
\label{app:generators}

To ensure that our evaluation is representative of the contemporary
text-to-image landscape rather than tied to a single architectural paradigm,
training regime or organizational lineage, we assemble 33 open-source and 9
proprietary generators. Our choice is guided by three criteria: generative
paradigm, model scale and release recency, and inference-time compute regime.

\emph{Coverage of generative paradigms.}
The latent diffusion paradigm is covered exhaustively through the Stable
Diffusion lineage (SD-1.4, SD-1.5, SD-2.1~\cite{rombach2022high},
SDXL~\cite{podell2024sdxl}, SD-3 Medium, SD-3.5 Medium and SD-3.5
Large~\cite{esser2024scaling}), through PixArt-$\alpha$~\cite{chen2024pixart_alpha}
and PixArt-$\Sigma$~\cite{chen2024pixart_sigma}, which isolate the effect of
DiT-style transformer backbones, and through Playground
v2.5~\cite{li2024playground}, Kolors~\cite{kwai_kolors_2024} and
Lumina-Image-2.0~\cite{qin2025lumina}, which probe alternative training recipes
on comparable architectures. Cascaded pixel-space diffusion is represented by
DeepFloyd~IF~\cite{deepfloyd_if_2023}, a counterpoint to latent approaches.
Flow-matching models are covered by the FLUX family (FLUX.1-dev,
FLUX.1-schnell~\cite{flux2024}, FLUX.1~Krea~[dev]~\cite{flux1kreadev2025},
FLUX.1-Kontext-dev~\cite{labs2025flux1kontextflowmatching}),
HiDream-I1-Full~\cite{hidreami1technicalreport}, Qwen-Image~\cite{wu2025qwen},
Chroma1-HD~\cite{rock2025chroma} and SRPO~\cite{shen2025directly}, which
together span the current open state of the art. Autoregressive and
visual-autoregressive synthesis is represented by
Janus-Pro-7B~\cite{chen2025janus}, Infinity-2B and
Infinity-8B~\cite{han2025infinity}, and by the unified multimodal models
OmniGen~\cite{xiao2025omnigen}, OmniGen2~\cite{wu2025omnigen2} and
Ovis-Image~\cite{wang2025ovis}.

We deliberately omit older GAN-based generators entirely. In preliminary experiments,
models such as DF-GAN~\cite{tao2022df} and GALIP~\cite{tao2023galip} proved
unable to follow prompts of the specificity our pipeline produces, and the
images they return are separated near-perfectly by most detectors in our pool.
Including them would inflate reported accuracy without measuring anything about
modern synthesis, and would reintroduce the content gap, which the prompt
design is meant to remove.

\emph{Temporal coverage and lineage diversity.}
Including SD-1.4 alongside SD-3.5 Large and the FLUX family lets us trace the
trajectory of a single open lineage across several generations under a
controlled architectural family, while the Kandinsky line
(Kandinsky-2~\cite{razzhigaev2023kandinsky},
Kandinsky-3.1~\cite{arkhipkin2023kandinsky},
Kandinsky-5~\cite{arkhipkin2025kandinsky}), Ovis-Image~\cite{wang2025ovis},
Z-Image-Turbo~\cite{cai2025z} and Qwen-Image~\cite{wu2025qwen} ensure that the
benchmark is not biased towards any single research ecosystem or pre-training
corpus. This organizational and geographic diversity matters because
text-to-image models inherit cultural, linguistic, and visual priors from their
data pipelines, and a benchmark drawn from a single lineage would systematically
under-report that variability. The proprietary tier --- GPT-Image-2,
Nano-Banana-2, Nano-Banana-Pro, Seedream-5-Lite, Imagen-4-Ultra, FLUX.2,
Grok-Imagine and Ideogram-v3 --- captures the closed-source frontier that, in
practice, most misuse in the wild is produced with.

\emph{Inference-time compute regimes.}
A growing fraction of practical deployments relies on few-step or single-step
samplers rather than the dozens of denoising steps assumed in the original
diffusion formulations. To make our evaluation robust to this shift, we include
distilled variants alongside their multi-step parents:
SDXL-Turbo~\cite{stabilityai_sdxl_turbo_2023} and
SDXL-Lightning~\cite{lin2024sdxl} against SDXL~\cite{podell2024sdxl}, SD-3.5
Large-Turbo against SD-3.5 Large~\cite{esser2024scaling}, FLUX.1-schnell against
FLUX.1-dev~\cite{flux2024}, as well as YOSO~\cite{luo2025you} and
Z-Image-Turbo~\cite{cai2025z} as recent few-step approaches. This lets us
measure detectability along the quality--latency frontier directly, rather than
inferring it through proxy metrics.

\subsubsection{Dataset Sampling and Split Construction}
\label{app:sampling}

Before sampling, the merged real--generated pool of $10.60$M images passes
through a second filtering stage applied \emph{symmetrically} to both classes.
Any criterion that real and generated images satisfy at systematically different
rates is a potential shortcut, so the thresholds on brightness, image
complexity, spectral characteristics, and no-reference quality are taken
from percentiles of the \emph{real} distribution, and the geometry bounds apply
identically to both. Generated images additionally pass a quality gate: a VLM
(InternVL3.5-38B) scores each of them for prompt relevancy, semantic coherence,
technical quality and scene complexity on a $1$--$10$ scale, and images below
$6.5$ on any axis are discarded as generation failures. This stage leaves
$4.2$M images ($1.3$M real, $2.9$M generated), the pool from which all splits
are drawn.

We use two different sampling strategies: (i) a density-based criterion for the
Expert subset, which is meant to hold the images that are statistically hardest
to attribute; and (ii) a stratified cluster-based criterion for the base splits,
which preserve the diversity of the source distribution.

\paragraph{Feature representation.}
Both criteria operate in a common low-dimensional space built from a heterogeneous per-image description. We collect $512$-D CLIP image embeddings; an $80$-D vector of per-class YOLO detection counts; a $21$-D multi-hot vector of VLM content-type tags (e.g., \emph{outdoor photo}, \emph{2d art}, \emph{night scene}, \emph{portrait}, etc.); and various scalar features spanning quality and aesthetics metrics, image structural complexity estimate, geometry, colorfulness, brightness, spatial information, face and OCR statistics, and frequency-, noise-, and forensics characteristics.

Estimating class densities directly in this space is impractical: the
dimensionality is high, the blocks live on wildly different scales, and naive
concatenation would let the CLIP embeddings dominate distance. We
therefore reduce the features in two steps. Each block is first mapped to a comparable
dimensionality using applicable transformations ($25$-D PCA for CLIP, $12$-D KL-NMF for YOLO detections, TF-IDF + SVD-decomposition for VLM tags) --- after which each block $g$ is rescaled by
$w_g = 1/\sqrt{d_g}$ so that all four contribute equally to the concatenated
distance. A single linear PCA then projects the concatenation to $12$
dimensions, retaining $97.3\%$ of the variance. The final projection is
deliberately linear and orthogonal: unlike a nonlinear embedding, it preserves
Euclidean distances up to the discarded variance, which is what makes the
subsequent nearest-neighbor density estimates meaningful. It is fit on a
class-balanced subsample, since fitting on the natural $1{:}2.5$
real-to-generated mix would tilt the principal axes towards the generated
distribution.

\paragraph{REALIS-Expert.}
To select the hardest images, we use a criterion that directly approximates the
local log density ratio $\log\!\big(p_{\text{real}}(x)/p_{\text{gen}}(x)\big)$
in the neighborhood of each image. The motivation comes from statistical decision theory: for a two-class problem the
Bayes-optimal decision boundary is exactly the level set on which the density
ratio equals its prior-corrected threshold, and classification error is maximal
there. Images whose local log ratio is close to $0$ therefore lie, in this
feature space, precisely in the region where \emph{any} classifier operating on
these features must be maximally uncertain. We note that the guarantee is
relative to the chosen feature space and not to pixel space: a detector reading
low-level synthesis fingerprints can in principle still separate these images.
What the criterion does provide is a generator-agnostic and detector-agnostic
way of removing the semantically and photometrically easy part of the
distribution.

Specifically, we build two HNSW indices~\cite{malkov2020hnsw} over the $12$-D projections of the
real and generated pools and estimate both log densities at every point with the
Loftsgaarden--Quesenberry $k$-nearest-neighbour estimator~\cite{loftsgaarden1965density}. Retaining only images with
$|\log \hat r(x)| \leq 0.5$ keeps $\sim\!12\%$ of the pool ($\sim\!500$k images) and by
itself lowers the average ROC-AUC of multiple tested pretrained detectors from $0.826$ to
$0.809$. Restricting to the strongest generators drops it further to $0.723$, and a
final VLM gate requiring a score of at least $8$ on every quality axis yields
$0.716$, against $0.871$ on the training split. REALIS-Expert is sampled
uniformly from what remains, giving $130$k images ($60$k real, $70$k generated)
across the $16$ hardest open-source and proprietary generators.

We further verified that the density signal is predictive of detector behavior,
and not merely of our own feature construction. Across the tested
detectors, $|\log \hat r(x)|$ correlates \emph{negatively} with per-image
detection error on generated images (PLCC reaching $-0.15$ for UFD and $-0.13$
for SAFE and RINE): the closer an image lies to the decision boundary in our
feature space, the more likely existing detectors are to misclassify it. The
effect is small in absolute terms --- as it must be, since our features contain
no low-level forensic information --- but it is consistent in sign across
architectures and highly significant at this sample size. 
We deliberately do \emph{not} use pretrained detector predictions as the
selection criterion for the hard subset, even though this would yield a subset
on which those detectors score far worse. Such a subset would encode the
specific blind spots of the current model generation and of the prior detection
datasets they were trained on, and a future detector that happens to fail
differently would be scored against difficulty defined by its predecessors. The
density criterion depends only on the data.

\paragraph{Base splits.}
For the train, validation, and test splits, the objective is the opposite: to
preserve the diversity of the source distribution across multiple feature
scales. We reuse the same $12$-D representation and partition it with the $K$-means clusterer. We chose $k = 20$ based on cluster-quality evaluations with silhouette, Davies--Bouldin~\cite{davies1979cluster} and Calinski--Harabasz~\cite{calinski1974dendrite} scores. Sampling then proceeds with equal quotas per cluster --- $55$k images per
cluster for training, and $2.5$k real plus $2.5$k generated per cluster for
validation and test --- which prevents the most popular content types of the web-image distribution (e.g., outdoor photographs of people) from crowding out sparser regions such as aerial or night scenes. Within each split, only images from that split's assigned generators are eligible. The resulting REALIS-Base contains $1.3$M images: $1.1$M train ($505$k real, $595$k generated), $100$k validation and $100$k test, both balanced $50/50$, over $30$ open-source generators.

\paragraph{Split disjointness.}
Splits are made disjoint at the level of \emph{prompts}, not images. Because each
real image induces one prompt and $3$--$4$ near-isosemantic generated
counterparts, splitting at the image level would place visually near-identical
content on both sides of a train/test boundary and inflate reported
generalization. Therefore, a real image and all of its generated counterparts
always live in exactly one split. 
The difficulty gradient the design targets is borne out empirically: averaged across pretrained detectors, mean ROC-AUC falls monotonically across the splits---from $0.871$ on Train and $0.854$ on Val splits, down to $0.836$/$0.716$ on the Test and Expert sets, respectively.

\section{Prompt Generation Instructions}
\label{app:prompt-instructions}
The prompts for image generation were produced using a Qwen3-30B-A3B LLM model prompted with the following instructions:
\begin{quote}
You rewrite long image captions into ONE short prompt with strict importance ordering.

INPUT CAPTION between << >>.

OUTPUT\break
- From one to three sentences of 50 words in total. Output ONLY the sentences.

ORDER \& RULES
\begin{enumerate}
\item  Begin with content type (if given/obvious) + exact counts + subjects + main action + main relation.
\item  Then add, as relevant:
\begin{enumerate}
        \item  For photos: scene/place/time $\rightarrow$ composition $\rightarrow$ lighting.
        \item  For screenshots/documents/charts/abstract/CGI: domain-specific layout/detail (e.g., centered layout, dark mode, A4 portrait, legend, north-up, scale bar) $\rightarrow$ optional style (vector/schematic/CGI/photoreal if explicitly stated).
\end{enumerate}
\item  If on-image text is mentioned in the caption, state it EXACTLY as it is, with its location.
\item  No invented details or quality words. If >50 words, drop from the tail in this order:
domain-neutral palette $\rightarrow$ camera $\rightarrow$ extra style $\rightarrow$ secondary layout.
Never drop content type, entities, counts, relations, or required text.
\item  Output sentences MUST be in English.
\end{enumerate}

Now produce the sentence.

<<CAPTION\_START>>\break
*caption*\break
<<CAPTION\_END>>
\end{quote}

\section{Generator Inventory Across Splits}
\label{app:generator-inventory}

Table~\ref{tab:generator_inventory} lists the generators of every split,
grouped by release year. Older models are concentrated in the training split,
validation and test add newer ones, and REALIS-Expert is formed by models
released from late 2024 onwards together with all proprietary generators; of
its open-source models, only HiDream-I1-Full also appears in a base split
(Test). Five generators (DeepFloyd IF, FLUX.1-Kontext-dev, Infinity-8B,
OmniGen2 and Ovis-Image) are present in all three base splits, so that
performance on seen and unseen generators can be compared within the same
evaluation split, while Playground v2.5 and Lumina-Image-2.0 are shared by
validation and test only.

\begin{table}[t]
\centering
\caption{Generators in each REALIS split, grouped by public release year of the evaluated model. Train, validation and test form REALIS-Base. $^\ast$Also in the training split, which allows seen and unseen generators to be compared on the same detector; $^\dagger$proprietary, accessed through public APIs and used only in REALIS-Expert. The last two rows give the number of generators and of images (real / generated) per split.}
\label{tab:generator_inventory}
\footnotesize
\setlength{\tabcolsep}{4pt}
\begin{tabularx}{\textwidth}{@{}l*{4}{>{\raggedright\arraybackslash}X}@{}}
\toprule
Year & \textbf{Train} & \textbf{Validation} & \textbf{Test} & \textbf{Expert} \\
\midrule
2022 & SD-1.4 \newline SD-1.5 \newline SD-2.1 & --- & --- & --- \\
\midrule
2023 & DeepFloyd IF \newline Kandinsky-2 \newline PixArt-$\alpha$ \newline SDXL \newline SDXL-Turbo & DeepFloyd IF$^\ast$ & DeepFloyd IF$^\ast$ & --- \\
\midrule
2024 & Infinity-2B \newline Kandinsky-3.1 \newline Kolors \newline OmniGen \newline PixArt-$\Sigma$ \newline SDXL-Lightning \newline YOSO-PixArt-512 & FLUX.1-schnell \newline Playground v2.5 \newline SD-3 Medium \newline Switti-1024 & FLUX.1-dev \newline Playground v2.5 \newline SD-3.5 Medium & SD-3.5 Large \newline SD-3.5 Large-Turbo \\
\midrule
2025 & FLUX.1-Kontext-dev \newline Infinity-8B \newline OmniGen2 \newline Ovis-Image & FLUX.1-Kontext-dev$^\ast$ \newline Infinity-8B$^\ast$ \newline OmniGen2$^\ast$ \newline Ovis-Image$^\ast$ \newline Lumina-Image-2.0 & FLUX.1-Kontext-dev$^\ast$ \newline Infinity-8B$^\ast$ \newline OmniGen2$^\ast$ \newline Ovis-Image$^\ast$ \newline HiDream-I1-Full \newline Lumina-Image-2.0 & FLUX.1 Krea [dev] \newline FLUX.1-dev SRPO \newline HiDream-I1-Full \newline Kandinsky-5 \newline Z-Image-Turbo \newline FLUX.2 [pro]$^\dagger$ \newline Imagen-4-Ultra$^\dagger$ \\
\midrule
2026 & --- & --- & --- & GPT-Image-2$^\dagger$ \newline Grok-Imagine$^\dagger$ \newline Nano-Banana-2$^\dagger$ \newline Qwen-Image-2.0$^\dagger$ \newline Recraft-V4$^\dagger$ \newline Seedream-5.0-Lite$^\dagger$ \newline Wan-2.7-Image-Pro$^\dagger$ \\
\midrule
Generators & 19 & 10 & 10 & 7 + 9$^\dagger$ \\
Images & 504.6k / 595.4k & 50k / 50k & 50k / 50k & 60k / 69.1k \\
\bottomrule
\end{tabularx}
\end{table}

\section{Complete List of Post-Processing Transformations}
\label{app:transformations}

Tables~\ref{tab:transformations} and \ref{tab:transformations2} list all transformations of the degradation
pipeline used to produce the distorted variants of the validation, test and
Expert splits (Section~\ref{sec:dataset_construction}). The $34$ core
transformations are organized into ten groups. For every image, 1-5 groups are drawn without replacement (each count with equal probability),
one transformation is selected uniformly within each group, and the selected
transformations are applied in sequence. Every transformation has five strength
levels, and the level of each selected transformation is drawn independently
from a discretized Gaussian over the level index; in the released splits,
levels~1--5 occur with frequencies of about $13$, $17$, $22$, $25$ and $23\%$,
respectively. After the chain, an invisible watermark produced by one of $12$
methods is embedded with probability $0.35$, the image is further downscaled by a
random factor in $[0.3, 0.8]$ with probability $0.75$, and the result is saved as near-lossless JPEG at quality $95$, with the full distortion chain logged for every image. %

To indicate the severity of each setting, the last two columns of
Table~\ref{tab:transformations} report PSNR and SSIM for every transformation
applied in isolation at its weakest and strongest level. The two metrics are
complementary: global photometric changes such as RGB shift, tone curves and
saturation changes lower PSNR to $15$--$20$\,dB at level~5 while SSIM stays at
or above $0.90$, whereas noise, glass blur and strong darkening degrade
structure the most (SSIM down to $0.13$ for shot noise). Neural codecs,
adversarial embedding attacks and WMForger remain at or above $29$\,dB and
$0.85$ SSIM even at level~5. Because transformations are chained and followed by
watermarking and downscaling, the cumulative degradation of a distorted image is
generally larger than these single-transformation values.

\section{Additional Detector Results}
\label{app:detector-results}
\begin{itemize}
\item \textbf{Conventional detectors:} AIDE~\cite{yan2025sanity},
BFREE~\cite{guillaro2025bfree}, DIRE~\cite{wang2023dire},
DistilDIRE~\cite{lim2024distildire}, DRCT~\cite{chen2024drct},
EFFORT~\cite{yan2025effort}, FatFormer~\cite{liu2024fatformer},
FreqNet~\cite{tan2024freqnet}, GRAG~\cite{gragnaniello2021grag},
GramNet~\cite{liu2020gramnet}, LASTED~\cite{wu2026lasted},
LGrad~\cite{tan2023lgrad}, MaskSim~\cite{li2024masksim},
NPR~\cite{tan2024npr}, ProDet~\cite{cheng2024prodet},
RINE~\cite{koutlis2025rine}, SAFE~\cite{li2025safe},
SPAI~\cite{karageorgiou2025spai}, UFD (UniFD)~\cite{ojha2023towards},
DMID and SelfCon.
\item \textbf{Zero-shot VLM detectors:} Qwen3-VL~\cite{bai2025qwen3vl}
(235B-A22B, 32B, 30B-A3B, 30B-A3B think, 8B, 8B think, 4B);
Qwen2.5-VL~\cite{bai2025qwen25vl} (32B, 7B, 3B);
Gemma-4~\cite{gemmateam2026gemma4} (31B, 12B, 4B, 2B);
InternVL3.5~\cite{wang2025internvl35} (30B-A3B, 14B, 8B, 4B).
\end{itemize}

This section extends the per-generator analysis of
Sections~\ref{sec:generator_generalization} and ~\ref{sec:content_statistics_performance}.

\subsection{Per-Generator Results}
\label{app:per-generator-results}

\begin{figure*}[tbp]
    \centerline{\includegraphics[width=0.99\textwidth]{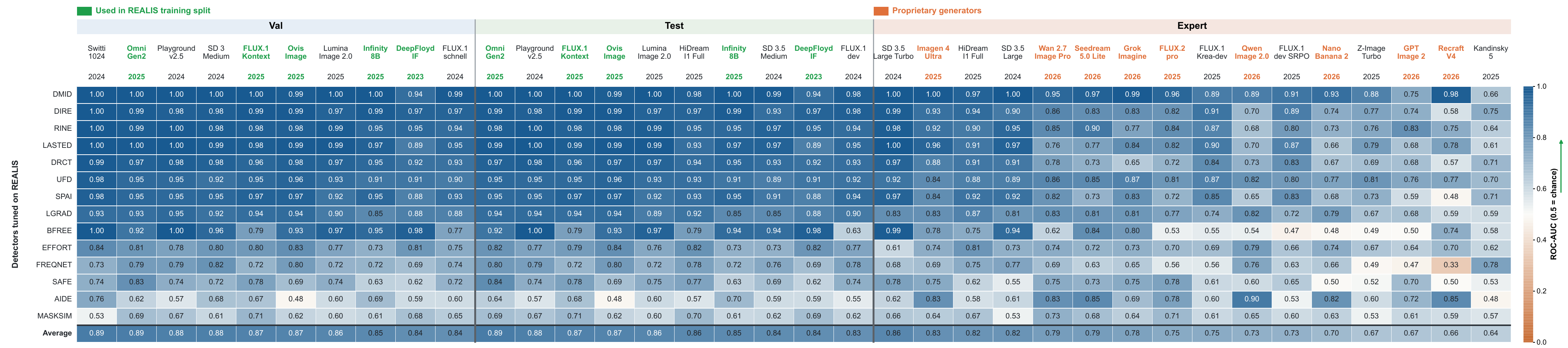}}
    \caption{ROC-AUC of the REALIS-tuned detectors on every generator of the
    validation, test and Expert splits. For each detector, the real images of
    all three splits are pooled into a single reference set ($160$k images),
    and each cell compares the images of one generator with this set. Green
    labels mark generators used in the REALIS training split, orange labels
    mark proprietary generators, and release years are given below the names.
    Columns are ordered within each split by decreasing average AUC, and the
    last row averages the detectors.}
\label{fig:app_auc_pooled}
\end{figure*}

\begin{figure*}[tbp]
    \centerline{\includegraphics[width=0.85\textwidth]{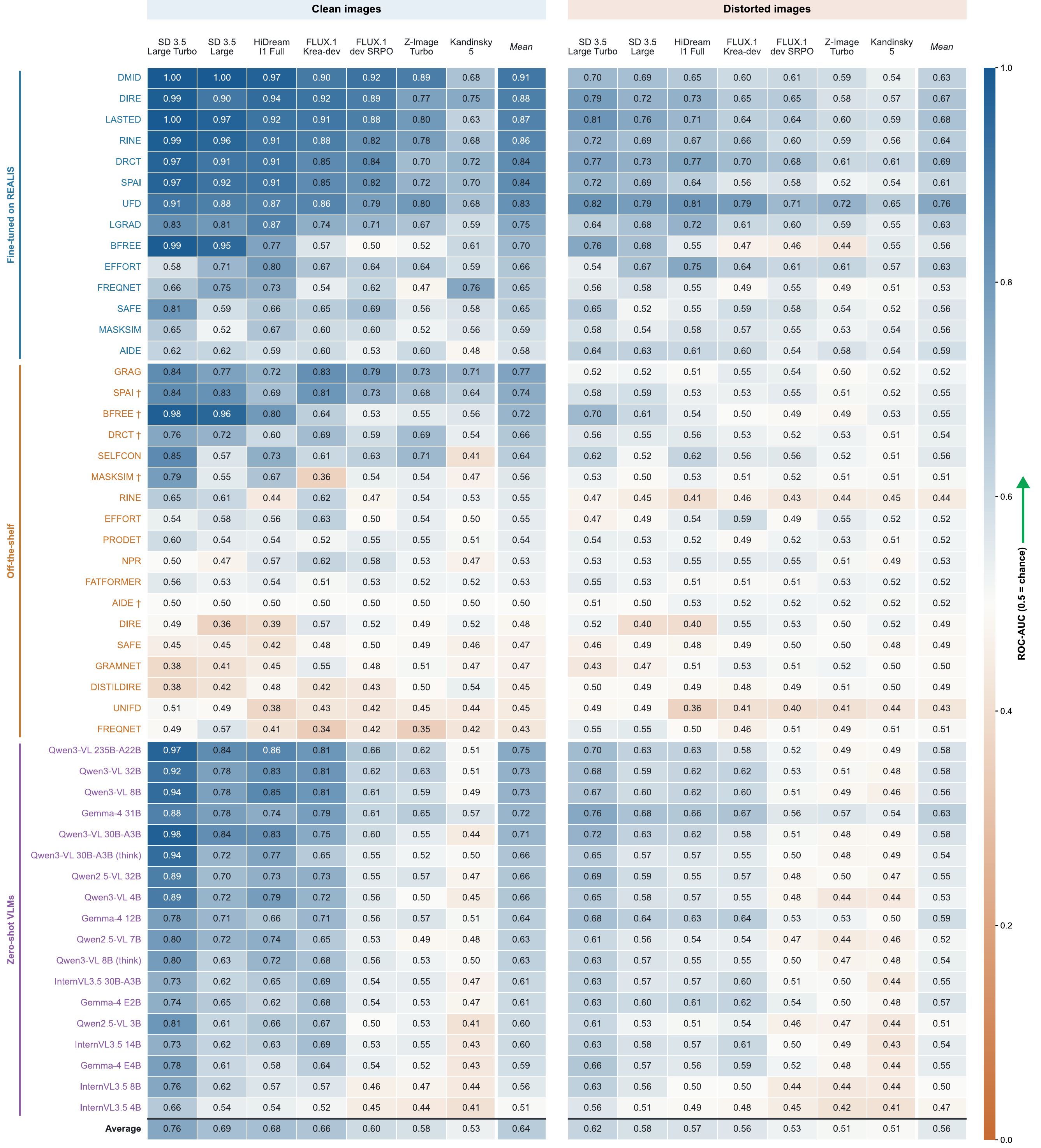}}
    \caption{Per-generator ROC-AUC for all tested detectors on the open-source part of REALIS-Expert,
    for clean images (left) and their distorted versions (right). Each cell
    compares the images of one generator with all real Expert images of the
    same version, and for each detector both halves use the same images;
    \emph{Mean} averages the seven generators. Rows are grouped into detectors
    fine-tuned on REALIS, off-the-shelf detectors and zero-shot VLMs, and are
    ordered by clean AUC within each group. $\dagger$:~the detector's output
    contains only binary decisions, so its AUC equals balanced accuracy.}
\label{fig:app_clean_distorted}
\end{figure*}

\textbf{Pooled real reference.} Figure~\ref{fig:app_auc_pooled} extends
Figure~\ref{fig:per_gen_results_tests} to the validation split. Since the real
images of Val, Test and Expert form one reference set for each detector,
differences between columns reflect only the generated images. Averaged over
detectors and generators, the AUC is $0.867$ on Val and $0.859$ on Test, but
$0.754$ for the open-source and $0.744$ for the proprietary Expert generators.
Generators used in training are not detected consistently better than held-out
ones: their mean AUC is $0.864$ versus $0.870$ on Val and $0.865$ versus
$0.853$ on Test. On Expert, generator averages range from $0.86$ for
SD-3.5 Large-Turbo to $0.64$ for Kandinsky-5. The proprietary models are spread
over this range, from Imagen-4-Ultra ($0.83$) to Recraft-V4 ($0.66$).

\textbf{Clean and distorted images.} Figure~\ref{fig:app_clean_distorted}
repeats the per-generator comparison on the distorted Expert images of
Section~\ref{sec:postprocessing_robustness}. Distortion lowers the AUC in every
detector--generator cell with both runs, from $0.79$ to $0.60$ on average. The
losses are largest for the generators that are easiest on clean images
(SD-3.5 Large-Turbo $0.93\rightarrow0.69$, SD-3.5 Large $0.87\rightarrow0.65$)
and smallest for Kandinsky-5 ($0.62\rightarrow0.53$), so the gap between the
easiest and the hardest generator halves, from $0.31$ to $0.16$. Averaged over
generators, the fine-tuned reference detectors fall from $0.87$ to $0.65$, the
zero-shot VLMs from $0.73$ to $0.59$, and the off-the-shelf detectors from
$0.72$ to $0.54$, close to chance. Across all detectors with both runs,
including weaker ones, $82\%$ of the cells decrease and $24\%$ of the distorted
cells fall below $0.5$.

\subsection{Detection at a Fixed Operating Threshold}
\label{app:fixed-threshold}

\begin{figure*}[tbp]
    \centerline{\includegraphics[width=0.99\textwidth]{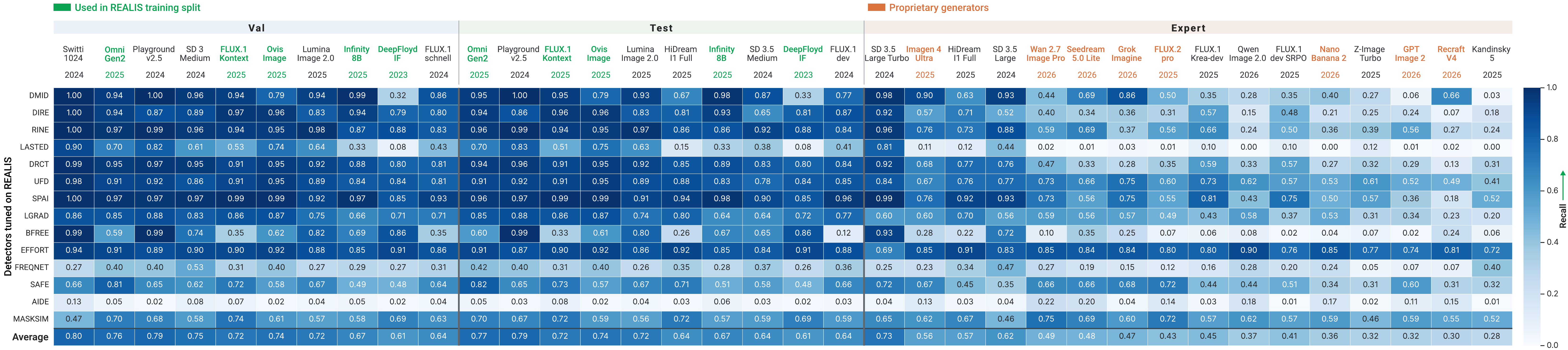}}
    \caption{Per-generator recall of REALIS-tuned detectors on the validation, Test, and Expert splits. Recall is the fraction of generated images assigned a score greater than $0.5$. Green labels identify generators represented in training, orange labels identify proprietary generators, and the bottom row averages the displayed detectors equally. Low recall on several Expert generators reveals limitations of the default operating threshold that are not fully captured by ROC-AUC; recall should be interpreted together with false-positive rates.}
\label{fig:recall_per_gen}
\end{figure*}

\begin{figure*}[tbp]
    \centerline{\includegraphics[width=0.99\textwidth]{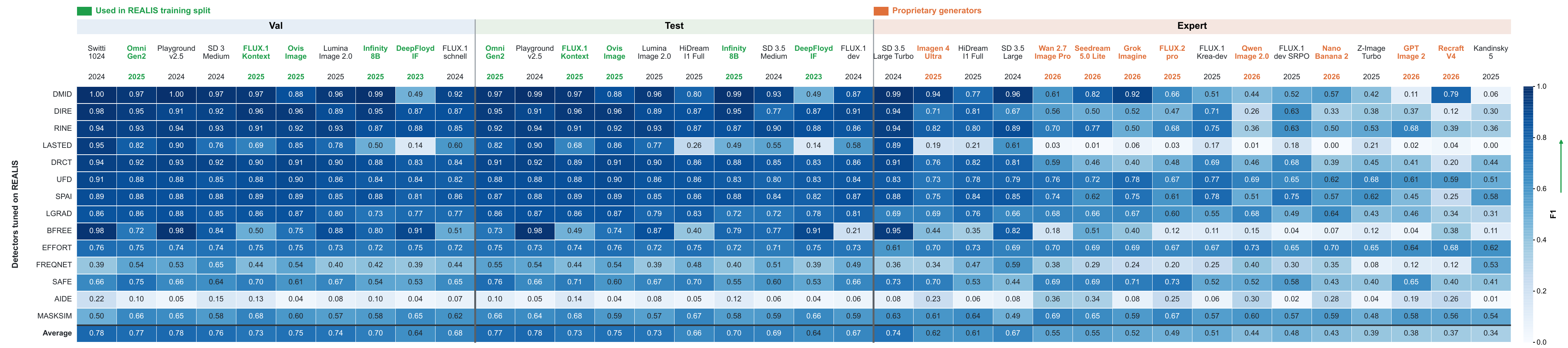}}
    \caption{F1 of the REALIS-tuned detectors at the default threshold of
    $0.5$, computed with real and generated images weighted equally (calibrated F1):
    $F_1 = 2R/(1 + R + \mathrm{FPR})$, where $R$ is the recall on the
    generator's images and FPR is the share of the split's real images with a
    score above $0.5$. Rows, columns and label colors as in
    Figure~\ref{fig:app_auc_pooled}.}
\label{fig:app_f1}
\end{figure*}

AUC measures how well scores rank generated above real images, independently of
a decision threshold. Figures~\ref{fig:recall_per_gen} and~\ref{fig:app_f1} instead
evaluate the detectors at the default threshold of $0.5$. Averaged over
detectors and generators, recall falls from $72.1\%$ on Val and $69.3\%$ on
Test to $44.7\%$ on Expert ($48.3\%$ for open-source and $41.9\%$ for
proprietary generators). The false-positive rate on real images averages
$16\%$ in every split, but differs strongly between detectors: DMID, AIDE and
LASTED flag less than $1\%$ of real images (LASTED less than $0.05\%$), whereas
MASKSIM flags $41$--$43\%$ and EFFORT $52$--$59\%$. A high AUC can therefore coincide with low
recall. LASTED, for example, reaches a pooled Expert AUC of $0.81$ but a recall
of only $12\%$, because it scores almost all images of both classes below the
threshold.

The calibrated F1 score~\cite{Siblini_2020} in Figure~\ref{fig:app_f1} combines recall and false positives in
one number per generator. Since a generator contributes between $1$k and $12$k
images, whereas each split has $50$k--$60$k real images, F1 is computed with both
classes weighted equally; otherwise it would mainly reflect the number of
images per generator. Mean F1 is $0.73$ on Val, $0.71$ on Test and $0.51$ on
Expert ($0.54$ for open-source and $0.48$ for proprietary generators). The
ranking of detectors by F1 differs markedly from the ranking by AUC (Spearman
correlation $0.35$ on Expert). DMID has the highest Expert AUC ($0.92$) but
only the fifth-highest F1 ($0.63$), while UFD has the highest F1 ($0.70$).
Because F1 ignores correctly rejected real images, it also favors detectors
that flag many images: EFFORT ranks second ($0.68$) despite a false-positive
rate of $59\%$ on Expert real images. F1 should therefore be read together with
the false-positive rate.

\section{Additional Analyses by Image Characteristics}
\label{app:image-characteristics}

\subsection{Generator-Balanced Statistical Sweeps}
\label{app:balanced-sweeps}

\begin{figure*}[tbp]
    \centerline{\includegraphics[width=0.99\textwidth]{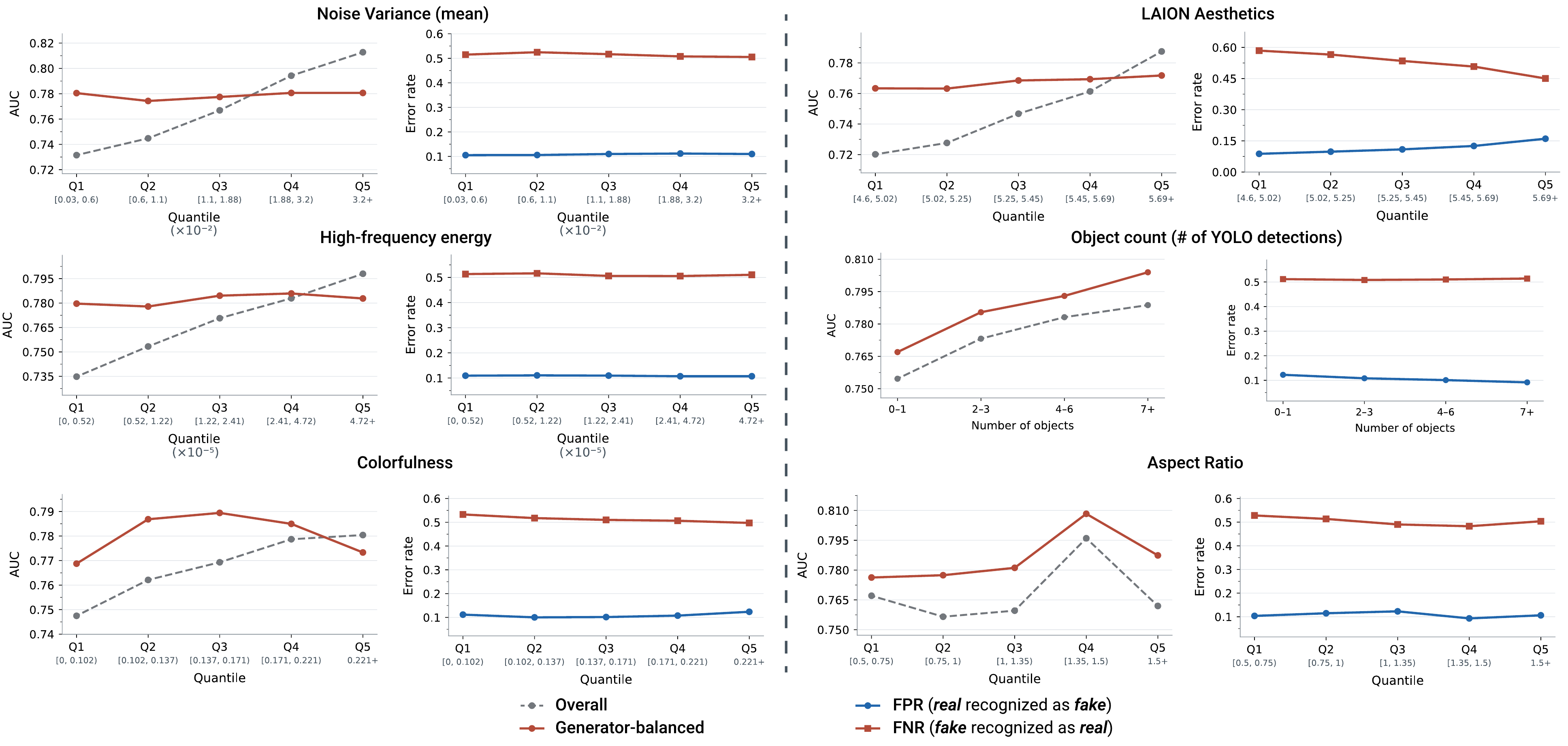}}
    \caption{Detection performance across quantile bins of six per-image
    statistics on the open-source part of REALIS-Expert (seven generators,
    120k images), averaged over the 16 reference detectors of
    Figure~\ref{fig:content_analysis}. For each statistic, the left panel shows
    the overall AUC, which pools all generated images of a bin, and the
    generator-balanced AUC, which averages per-generator AUCs against the real
    images of the same bin; the right panel shows the FPR on real images and
    the generator-balanced FNR at a score threshold of $0.5$. Noise variance is the mean over
    $8{\times}8$ patches of the variance of the Laplacian response; high-frequency
    energy is the share of spectral power above half the Nyquist frequency;
    colorfulness follows Hasler and S\"usstrunk\cite{cf}; object count is the number of
    YOLO11x detections with confidence above $0.5$.}
\label{fig:app_balanced_sweeps}
\end{figure*}

Figure~\ref{fig:app_balanced_sweeps} extends the statistic sweeps of
Figure~\ref{fig:content_analysis}(b) to six further image statistics, using the
same protocol: quantile bins are computed on real and generated images
together, and the real reference of each AUC comes from the same bin. Since
every generator has its own characteristic noise, spectral and aesthetic
profile, the mixture of generators changes from bin to bin, and the overall AUC
inherits this change; the generator-balanced AUC removes it by weighting the
generators in each bin equally.

\textbf{Pooled trends explained by generator composition.} For estimated noise
variance, the overall AUC rises from $0.73$ in the lowest to $0.81$ in the
highest quintile, whereas the generator-balanced AUC stays between $0.774$ and
$0.781$. High-frequency energy behaves in the same way ($0.74$ to $0.80$
overall, $0.778$--$0.786$ balanced), and for the LAION Aesthetics score the
range shrinks from $0.72$--$0.79$ to $0.76$--$0.77$. For these three statistics
the balanced spread retains only $8$--$13\%$ of the pooled spread, and for
noise variance and high-frequency energy the error rates are flat as well.
Aesthetics is the exception: although its balanced AUC barely changes, the FPR
rises from $8.8\%$ to $16.0\%$ and the FNR falls from $58.4\%$ to $45.0\%$
between the lowest and highest quintile. More aesthetic images of both classes
thus receive higher scores, a shift that changes the behavior at a fixed
threshold without improving the ranking.

\textbf{Associations that persist.} The object count keeps its association
after balancing: the generator-balanced AUC increases from $0.767$ for images
with at most one confident detection to $0.804$ for images with seven or more
objects. This increase is accompanied by fewer false positives (FPR from
$12.2\%$ to $9.2\%$), while the FNR stays near $51\%$. Colorfulness shows a weaker, non-monotone association: the balanced AUC is
$0.769$ for the least colorful and $0.773$ for the most colorful quintile, but
$0.785$--$0.789$ in between, and the FPR increases in the most colorful bin.

\textbf{A limitation of balancing.} Both AUC curves for aspect ratio peak for
landscape images with ratios in $[1.35, 1.5)$ (balanced AUC $0.808$, compared
with $0.776$--$0.787$ elsewhere). Kandinsky~5, the most difficult Expert
generator, produces no images in this range, so this bin's balanced AUC
averages six rather than seven generators. Repeating the sweep without
Kandinsky~5 removes the peak (balanced AUC $0.800$--$0.814$ in all bins).
Generator balancing equalizes the weight of the generators present in a bin,
but cannot compensate for a generator that is absent from it; associations with
geometry should therefore be read together with the output sizes each
generator supports. In the other five statistics, every generator contributes
at least $140$ images to every bin.

\subsection{Generation Quality and Generator Profiles}
\label{app:generator-quality}

\begin{figure*}[tbp]
    \centerline{\includegraphics[width=0.99\textwidth]{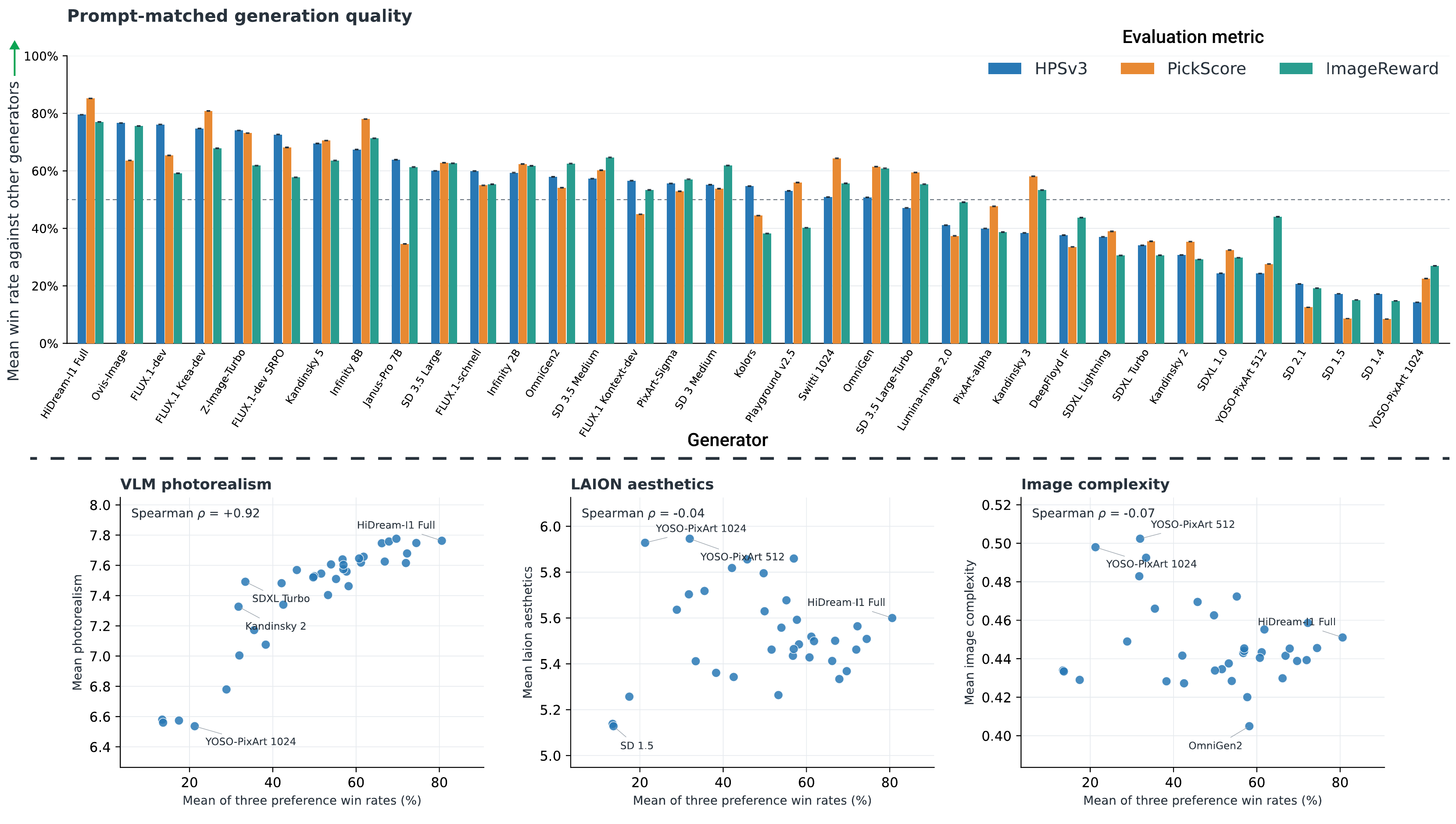}}
    \caption{Generation quality of the open-source generators, measured on
    all $8.39$M generated images before filtering. Top: mean prompt-matched win
    rate of each generator against the other $34$ under three automated
    preference models. For a pair of generators, the win rate is the share of
    common prompts on which the generator's image receives the higher score,
    with ties counted as one half; each bar averages these pairwise win rates
    with equal weight over opponents. Generators are ordered by HPSv3, the
    dashed line marks $50\%$, and error bars show $95\%$ prompt-cluster
    confidence intervals (at most $\pm0.16$ percentage points). Bottom:
    generator means of VLM photorealism score, LAION Aesthetics and image
    complexity (as measured by IC9600~\cite{ic_metric} model) against the average of the three win rates, one point per
    generator.}
\label{fig:generation_quality}
\end{figure*}

\begin{figure*}[tbp]
    \centerline{\includegraphics[width=0.99\textwidth]{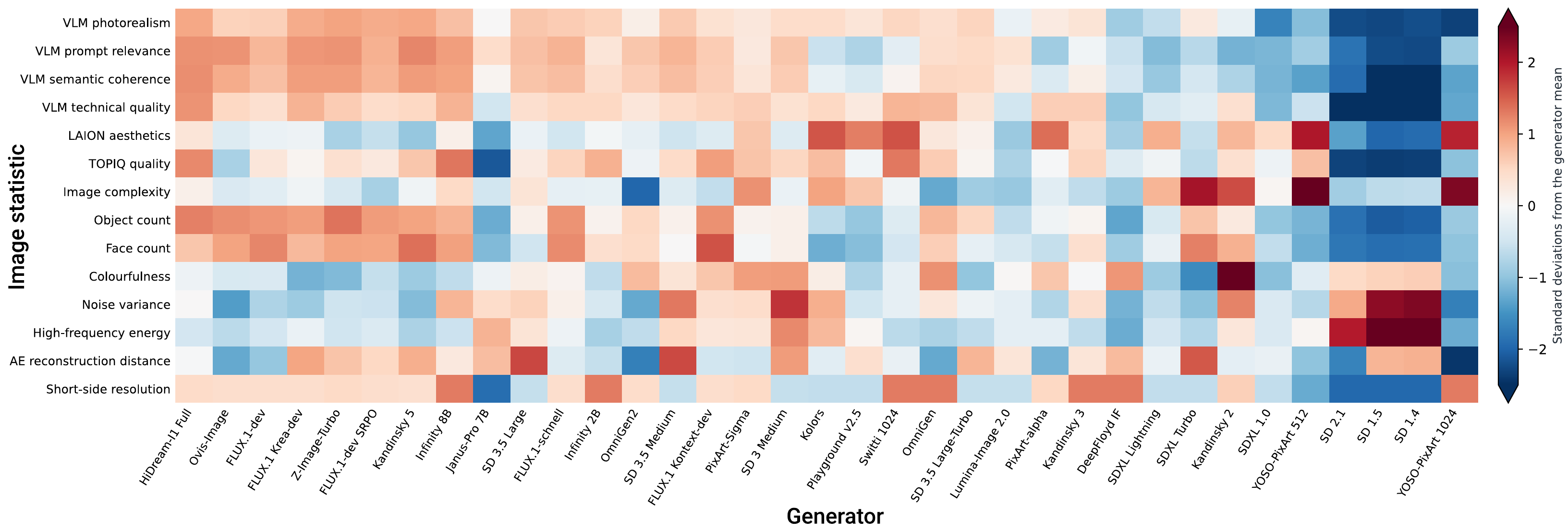}}
    \caption{Image-statistic profiles of the open-source generators on the
    same unfiltered pool. Each cell is a generator's mean of a statistic,
    standardized across the generator means and clipped at $\pm2.5$; color
    therefore compares generators within a row and does not indicate better or
    worse quality. Generators are ordered by HPSv3 win rate, as in
    Figure~\ref{fig:generation_quality}. VLM ratings are on a $1$--$10$ scale
    and come from the scoring step of the VLM quality gate
    (Section~\ref{sec:dataset_construction}); TOPIQ is a no-reference quality
    score; image complexity is the IC9600 estimate; AE reconstruction distance
    is the LPIPS distance used by the reconstruction filter; object count,
    colorfulness, noise variance and high-frequency energy are defined in
    Figure~\ref{fig:app_balanced_sweeps}.}
\label{fig:generator_statistics}
\end{figure*}

Figures~\ref{fig:generation_quality} and~\ref{fig:generator_statistics}
characterize the open-source generators themselves, without reference to any
detector. Both are computed on all $8.39$M images generated for REALIS before
filtering, and therefore describe each generator with the resolution and
inference settings used in our pipeline, not architectures compared under
identical settings.

\textbf{Prompt-matched quality.} Because each prompt was rendered by up to
three of these generators, quality can be compared on identical prompts. We
score all images with three automated preference models, HPSv3, PickScore and
ImageReward, and compute for every pair of generators the share of common
prompts (about $13$k per pair) on which one image scores higher. HiDream-I1-Full
wins most often under all three metrics ($79.5$, $85.2$ and $77.0\%$), whereas
SD-1.4, SD-1.5, SD-2.1 and YOSO-PixArt-1024 win at most $27\%$ of their
comparisons under any metric. The three rankings agree closely overall
(pairwise Spearman correlation $0.84$--$0.88$), but individual generators can
move far: Janus-Pro-7B ranks 9th under HPSv3 but 28th under PickScore, and
Switti-1024 ranks 8th under PickScore but 20th under HPSv3. As the confidence
intervals are this narrow, such differences reflect what the scoring models
reward rather than prompt sampling, and we report all three metrics instead of
a single ranking. Averaged over the three metrics and over the generators of
each split, the win rate rises from $42.7\%$ in the training split to $55.5\%$
in validation, $59.3\%$ in test and $67.8\%$ for the seven open-source Expert
generators, in line with the assignment of generators to splits by generation
quality (Section~\ref{sec:dataset_construction}). Within Expert, the
lowest-ranked generator, SD-3.5 Large-Turbo ($54.0\%$), is also the easiest to
detect (category-averaged AUC $0.93$ in Figure~\ref{fig:content_analysis}(a)).
However, the most difficult one, Kandinsky-5 (AUC $0.61$), ranks only fourth of
seven ($67.9\%$), and the top-ranked HiDream-I1-Full is comparatively easy to
detect (AUC $0.83$). Preference-based quality is therefore not a direct proxy
for detection difficulty.

\textbf{Generator profiles.} Across generators, the mean win rate agrees with
the VLM ratings (Spearman correlation $0.92$ with photorealism and
$0.93$--$0.95$ with prompt relevance and semantic coherence) and, more weakly,
with the numbers of detected objects ($0.78$) and faces ($0.64$). Aesthetic and
complexity scores capture something else: their correlations with the win rate
are $-0.04$ and $-0.07$, and the two YOSO-PixArt models, 29th and 32nd of 35 by
win rate, have the highest mean LAION aesthetic and complexity scores of all
generators. The oldest models, SD-1.4, SD-1.5 and SD-2.1, combine the lowest
VLM ratings and the smallest outputs with the highest high-frequency energy
($2.0$ to $3.0$ standard deviations above the generator average). Because
generators differ this much, the filters that are applied identically to both
classes still remove different shares of each generator's images. The
reconstruction filter discards images that the autoencoder reconstructs
unusually well, and it removes $67\%$ of all YOSO-PixArt-1024 images, the
generator with the lowest mean reconstruction distance, compared with at most
$35\%$ for any other generator.%

\section{Post-Processing Sensitivity by Transformation and Strength}
\label{app:postprocessing_analysis}

\begin{figure*}[t]
    \centering
    \includegraphics[width=\textwidth]{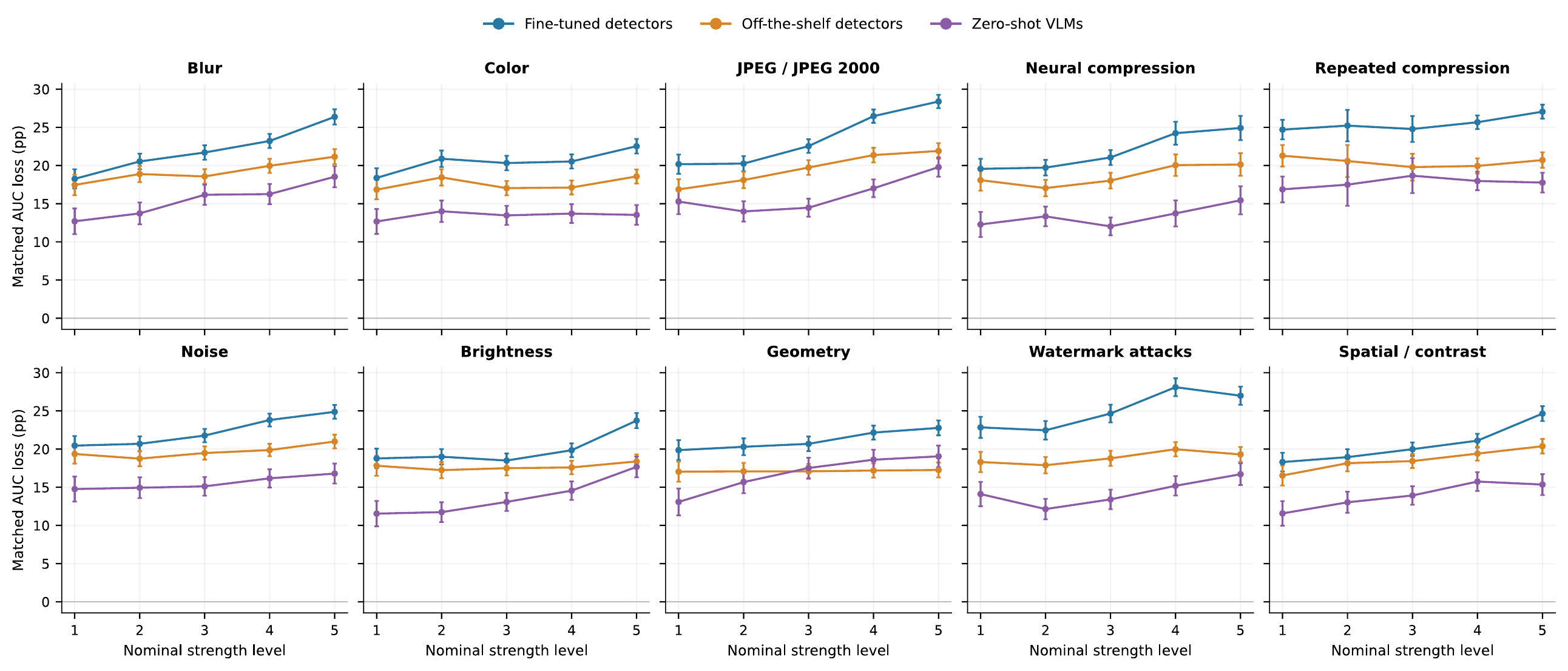}
    \caption{Matched AUC loss on Expert versus nominal transformation
    strength, conditioned on distortion-group membership. Curves average
    the fixed reference detectors within each detector group; error bars
    are pointwise $95\%$ prompt-clustered intervals. Each clean reference
    uses the same images as its distorted counterpart. Only uniquely
    recoverable levels are included. Operation mixtures and other
    processing stages can differ across points.}
    \label{fig:postprocessing_group_severity}
\end{figure*}

\textbf{Evaluation details.}
The Expert-split analyses provided in Section \ref{sec:postprocessing_robustness} use real images and images from the seven open-source
generators; proprietary generators are excluded. For each
distortion-containing subset in Figure \ref{fig:postprocessing_overview}a, we compute clean and distorted AUC on
identical images, average the seven generator-versus-real AUCs equally,
and then average over the reference detectors. Figure \ref{fig:postprocessing_overview}a employs 14 representative detectors of various types selected by their performance on non-distorted splits: 6 fine-tuned detectors, 4 off-the-shelf detectors,
and 4 VLMs. Figure \ref{fig:postprocessing_overview}b includes all 50 available paired
detectors. Positive matched AUC loss in Figure \ref{fig:postprocessing_group_severity} denotes clean minus distorted AUC. Error bars
are pointwise $95\%$ intervals obtained from paired AUC influence
functions clustered by image stem, conditional on the chosen detectors
and generators. Some off-the-shelf exports contain binary predictions,
so their AUC reflects a single operating point rather than a continuous
score ranking.

\begin{figure*}[t]
    \centering
    \includegraphics[width=0.95\textwidth]{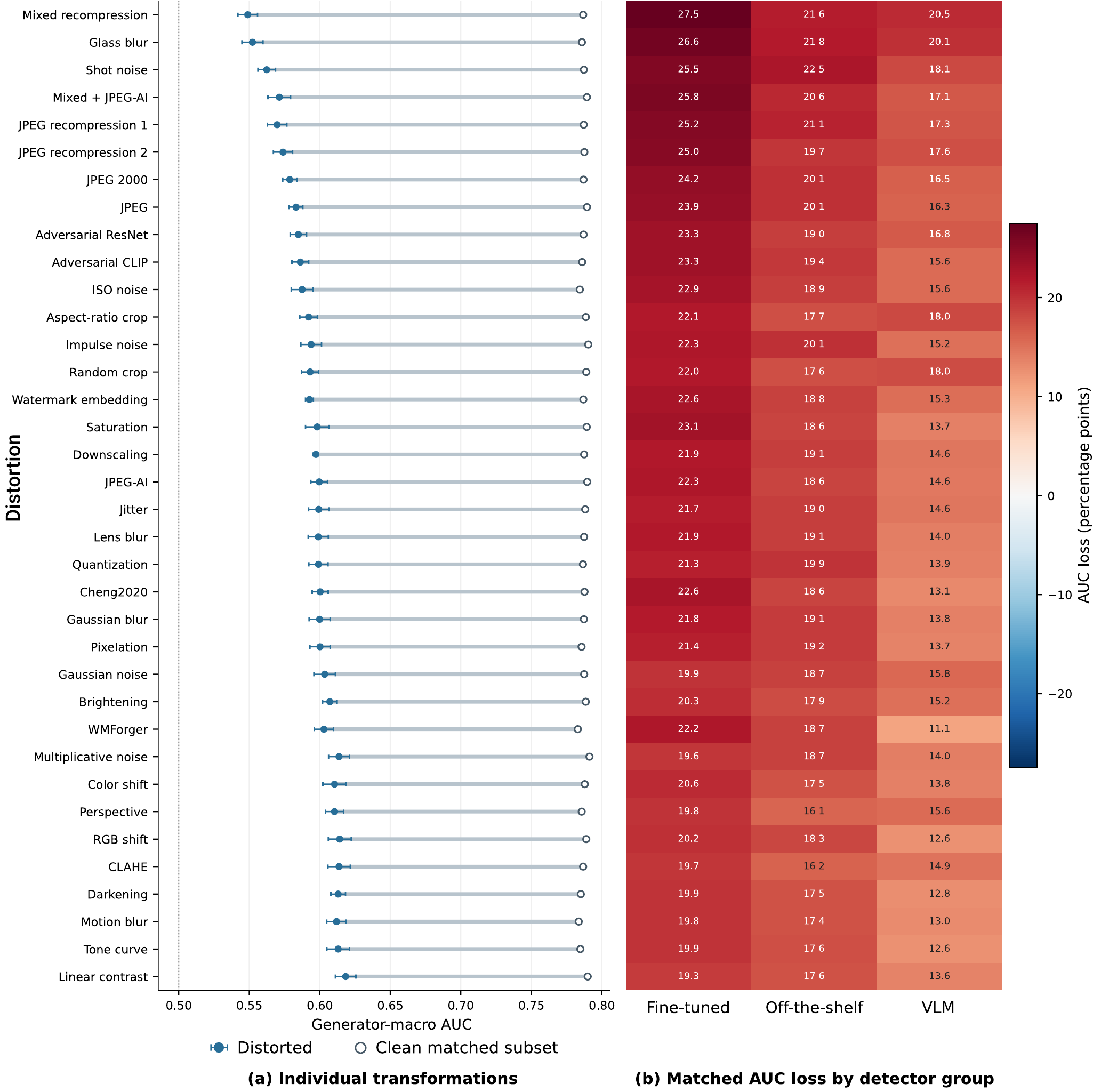}
    \caption{Individual transformations within the Expert post-processing
    pipeline. \textbf{(a)} Generator-macro AUC on each transformation-containing
    subset and its matched clean counterpart, averaged over the 14
    reference detectors. \textbf{(b)} Matched AUC loss, in percentage
    points, averaged separately within each detector group. Rows are
    ordered by mean loss. Error bars in (a) are pointwise $95\%$
    prompt-clustered intervals for distorted AUC. The subsets overlap
    and can include additional transformations.}
    \label{fig:postprocessing_individual}
\end{figure*}

\textbf{Individual transformations reveal variation within broad groups.}
Figure~\ref{fig:postprocessing_individual} shows nearly constant clean
reference AUC across the transformation-containing subsets, but
substantially different performance after processing. Mixed
recompression, glass blur, and shot noise give the lowest mean distorted
AUCs, whereas tone-curve, linear-contrast, and brightness changes are
associated with smaller losses. The distinction also holds within
broad categories: glass blur is more damaging than Gaussian or motion
blur, and shot noise more damaging than Gaussian or multiplicative
noise in these pipeline outputs. Broad labels such as ``blur'' or
``noise'' therefore conceal meaningful variation between operations.
The detector-group heatmap shows a broadly shared ordering, with
model-group differences in its magnitude. Even the lower-loss rows
contain other processing stages, so they should not be interpreted as
measurements of harmless individual transformations.

\textbf{Sensitivity to severity depends on the transformation group.}
Figure~\ref{fig:postprocessing_group_severity} conditions on both group
membership and the recorded nominal strength level. Blur and JPEG
compression show pronounced increases in AUC loss with strength across
all three detector groups. Color changes have flatter, less monotonic
profiles, while repeated compression incurs a large loss throughout
the range. Geometry is also detector-dependent: VLM losses increase
with strength, whereas the off-the-shelf mean changes relatively little.
The levels are operation-specific settings, not a common perceptual
distance, and each point samples different images and co-occurring
transformations. Only uniquely recoverable levels are used: repeated
parameter values that cannot distinguish two nominal levels are
excluded from these curves. These considerations are particularly
relevant to neural and repeated compression, where the operation mix
can vary between the plotted levels.

\begin{figure*}[t]
    \centering

    \includegraphics[width=\textwidth]{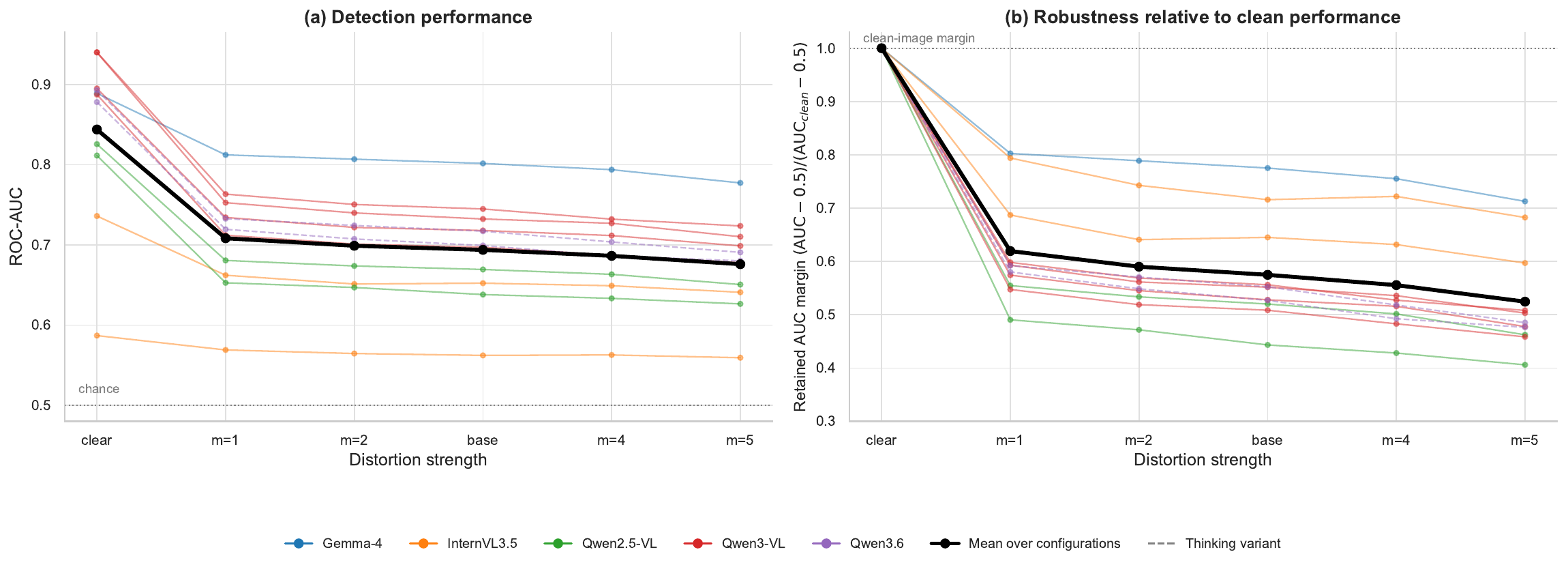}
    \caption{VLM sensitivity to the pipeline strength parameter on a
    separate validation sample. \textbf{(a)} Pooled ROC-AUC on clean
    images and pipeline outputs as $m$ varies; ``base'' denotes $m=3$.
    \textbf{(b)} Fraction of each configuration's clean AUC margin
    above chance retained after processing. Thin curves show individual
    configurations, dashed curves indicate thinking variants, and thick
    black curves average the configurations shown. The strongest setting
    has fewer available predictions; see the evaluation details in
    Appendix~\ref{app:postprocessing_analysis}.}
    \label{fig:robustness_strength}
\end{figure*}

\textbf{Enabling the pipeline causes a larger initial drop than subsequent
strength increments.}
Figure~\ref{fig:robustness_strength} complements the Expert analysis
with a separate validation sample of 5,000 real and 5,000 generated
images spanning ten generators. The pipeline parameter $m$ changes the
sampling distribution over transformation strengths, while its other
settings remain fixed; ``base'' denotes $m=3$. Thus, $m=1$ still enables
a compound pipeline, rather than applying only level-1 transformations.
For the configurations shown, mean pooled ROC-AUC falls from about
$0.84$ on clean images to $0.71$ at the lowest pipeline setting, then
to approximately $0.68$ at the highest. Panel~(b) normalizes each
configuration by its clean AUC margin above chance,
$({\rm AUC}_{m}-0.5)/({\rm AUC}_{\rm clean}-0.5)$; the mean retained
margin is roughly one half at the strongest setting. Gemma-4~31B
retains more of its margin than the displayed Qwen3-VL configurations,
consistent with their different relative losses on Expert. This
auxiliary experiment uses pooled AUC and a different generator and
detector cohort from the Expert analysis. Its strongest setting also
has incomplete prediction coverage, so the endpoints describe
available-case trends rather than a fully matched comparison across
all settings.%

\begin{table}[p]
\centering
\caption{Transformations of the REALIS degradation pipeline, grouped as sampled (Appendix~\ref{app:transformations}); parameters are given from level~1 to level~5. PSNR (dB, RGB) and SSIM (luma): median fidelity to the unprocessed image when the transformation is applied \emph{alone} at level~1 and at level~5, over 600 images (300 real, 300 generated).}
\label{tab:transformations}
\footnotesize
\setlength{\tabcolsep}{4pt}
\begin{tabularx}{\textwidth}{@{}l>{\raggedright\arraybackslash}Xcc@{}}
\toprule
Transformation & Description (level 1 to level 5) & PSNR, dB & SSIM \\
\midrule
\multicolumn{4}{@{}l}{\emph{Blur}} \\
Gaussian blur & Isotropic Gaussian filter, $\sigma$ = 0.1--5\,px. & $\infty$--24.5 & 1.00--0.77 \\
Lens blur & Uniform disk (defocus) kernel, radius 1--8\,px. & 38.6--24.0 & 0.99--0.76 \\
Motion blur & Linear motion kernel at a random angle, length 1--10\,px. & $\infty$--28.4 & 1.00--0.88 \\
Glass blur & Gaussian blur ($\sigma{=}0.5$) with local pixel swaps, offset 1--6\,px. & 25.2--19.5 & 0.78--0.50 \\
\addlinespace[2pt]
\multicolumn{4}{@{}l}{\emph{Color distortion}} \\
Color shift & Edge-weighted misregistration of the green channel, 1--12\,px. & 39.4--30.5 & 0.99--0.97 \\
RGB shift & Random per-channel offset of up to $\pm$10--$\pm$50 (of 255). & 34.0--19.7 & 1.00--0.98 \\
Saturation & Saturation scaled by 0.4 to $-0.4$ (${\le}\,0$: gray or inverted chroma). & 23.2--15.4 & 0.99--0.90 \\
Tone curve & Random per-channel spline tone curve, scale 0.05--0.4. & 32.2--19.4 & 1.00--0.97 \\
CLAHE & Contrast-limited adaptive histogram equalization, clip limit 1--6.5. & 22.5--16.3 & 0.91--0.72 \\
\addlinespace[2pt]
\multicolumn{4}{@{}l}{\emph{Algorithmic compression}} \\
JPEG & JPEG compression, quality factor 45--7. & 36.9--28.0 & 0.97--0.85 \\
JPEG~2000 & JPEG~2000 wavelet compression, ratio 16:1--170:1. & 40.5--28.6 & 0.98--0.85 \\
\addlinespace[2pt]
\multicolumn{4}{@{}l}{\emph{Neural compression}} \\
JPEG~AI & JPEG~AI~\cite{jpeg_ai_standard} learned codec (v7), four decreasing rates (levels 4, 5 equal). & 38.6--30.0 & 0.99--0.92 \\
Cheng2020 & Learned codec of ~\cite{cheng2020learned}\ (anchor), quality 5--1. & 37.1--31.7 & 0.96--0.91 \\
\addlinespace[2pt]
\multicolumn{4}{@{}l}{\emph{Multiple and mixed recompression}} \\
JPEG recompression 1 & 2--5 JPEG round trips, each with quality drawn from $[7, 43]$. & 31.8--29.0 & 0.92--0.88 \\
JPEG recompression 2 & Three JPEG round trips, quality range $[15, 45]$ to $[7, 25]$. & 32.4--29.1 & 0.93--0.88 \\
Mixed recompression & 2--5 round trips, each JPEG ($q\in[7,43]$) or JPEG~2000 (16--170:1). & 31.0--28.4 & 0.90--0.85 \\
Mixed + JPEG~AI & 2--5 round trips, each JPEG ($q\in[7,43]$) or JPEG~AI (random rate). & 30.9--27.5 & 0.93--0.89 \\
\addlinespace[2pt]
\multicolumn{4}{@{}l}{\emph{Noise}} \\
Gaussian noise & Additive Gaussian noise, $\sigma \approx 8$--18 (of 255). & 30.3--23.6 & 0.86--0.62 \\
Multiplicative noise & Speckle noise $x(1+n)$, $n\sim\mathcal{N}(0,v)$, $v$ = 0.001--0.035. & 36.5--21.6 & 0.96--0.59 \\
Impulse noise & Salt-and-pepper noise on 0.1--2\% of pixel values. & 35.1--22.1 & 0.98--0.67 \\
ISO noise & Camera-sensor luma and chroma noise, intensity 0.2--0.3 to 0.5--0.75. & 29.2--17.5 & 0.78--0.36 \\
Shot noise & Poisson (photon) noise, scale 0.025--0.075 to 0.2--0.3. & 17.8--10.0 & 0.36--0.13 \\
\addlinespace[2pt]
\multicolumn{4}{@{}l}{\emph{Brightness change}} \\
Brightening & Spline curve raising mid-tones (RGB and CIELAB), amount 0.1--1.1. & 29.1--9.3 & 0.99--0.66 \\
Darkening & Spline curve lowering mid-tones, amount 0.05--0.8. & 33.9--11.0 & 1.00--0.45 \\
\addlinespace[2pt]
\multicolumn{4}{@{}l}{\emph{Geometric}} \\
Random crop & Random crop keeping 80--40\% of each side. & --- & --- \\
Aspect-ratio crop & Random crop, height 80--40\%, aspect ratio drawn from $[0.5, 2]$. & --- & --- \\
Perspective & Random perspective warp, corner offset scale 2.5--7.5\% to 30--40\%. & --- & --- \\
\addlinespace[2pt]
\bottomrule
\end{tabularx}
\vspace{3pt}
\parbox{\textwidth}{List continues in Table \ref{tab:transformations2}.}
\end{table}

\begin{table}[p]
\centering
\caption{Continuation of Table \ref{tab:transformations} describing transformations used in REALIS.}
\label{tab:transformations2}
\footnotesize
\setlength{\tabcolsep}{4pt}
\begin{tabularx}{\textwidth}{@{}l>{\raggedright\arraybackslash}Xcc@{}}
\toprule
Transformation & Description (level 1 to level 5) & PSNR, dB & SSIM \\
\midrule
\multicolumn{4}{@{}l}{\emph{Adversarial and watermark-removal attacks}} \\
Adversarial CLIP & PGD (100 steps) on the CLIP ViT-B/32 embedding, $\ell_\infty$ 4--12/255; method from \cite{an2024waves}. & 39.0--30.5 & 0.98--0.89 \\
Adversarial ResNet & Same PGD attack on a ResNet-18 embedding, $\ell_\infty$ 4--12/255; method from \cite{an2024waves}. & 38.3--30.2 & 0.97--0.87 \\
WMForger & Watermark removal method~\cite{soucek2025wmforger} with a learned preference model, step 0.05--1. & 36.4--29.1 & 0.97--0.85 \\
\addlinespace[2pt]
\multicolumn{4}{@{}l}{\emph{Spatial distortion}} \\
Pixelation & Nearest-neighbor down- and upsampling by 0.89--0.29. & 25.7--21.8 & 0.85--0.72 \\
Jitter & Random per-pixel displacement (5 iterations), std.\ 0.05--1\,px. & 43.0--25.3$^\dagger$ & 0.99--0.80 \\
Quantization & Uniform quantization to 20--7 intensity levels. & 31.5--22.0 & 0.97--0.88 \\
Linear contrast & Spline contrast curve, offset 0, 0.15, $-0.4$, 0.3, $-0.6$ (${<}\,0$ lowers contrast). & 27.8--21.8$^\dagger$ & 0.97--0.93$^\dagger$ \\
\addlinespace[2pt]
\multicolumn{4}{@{}l}{\emph{Applied after the chain}} \\
Watermark embedding & Invisible watermark by one of 12 methods\textsuperscript{\S}; $p{=}0.35$. & --- & --- \\
Downscaling & Resize by a factor drawn from $[0.3, 0.8]$; $p{=}0.75$. & --- & --- \\
\bottomrule
\end{tabularx}
\vspace{3pt}
\parbox{\textwidth}{\textsuperscript{\S}Employed methods: ARWGAN~(\cite{arwgan}), DCT-Marker~(\cite{dct_wm}), DWSF~(\cite{dwsf}), FIN~(\cite{fin}), MaskWM~(\cite{maskwm}), MBRS~(\cite{mbrs}), PIMoG~(\cite{pimog}), PixelSeal~(\cite{pixelseal}), SSHidden~(\cite{zhu2018hidden}), StegaStamp~(\cite{tancik2020stegastamp}), SyncSeal~(\cite{syncseal}), TrustMark~(\cite{bui2023trustmark}). Implementations taken from WIBE framework (\cite{yakushev2025wibe}).}
\end{table}

\end{document}